\documentclass{article}
\usepackage{jfr2018stylee}
\usepackage{graphicx}
\usepackage{apalike}
\usepackage{setspace}
\usepackage{subfig}
\usepackage{url}
\usepackage{amsmath}
\usepackage{amsfonts}
\usepackage{amssymb}
\usepackage{multirow}
\usepackage{lineno}

\usepackage{hyperref}
\graphicspath{{figures/}}

\newcommand{\specialcell}[2][c]{%
	\begin{tabular}[#1]{@{}c@{}}#2\end{tabular}}

\title{A Comparative Study of Fruit Detection and Counting Methods for Yield Mapping in Apple Orchards}

\author{
Nicolai H{\"a}ni\ \\
Department of Computer Science \& Engineering\\
University of Minnesota\\
Minneapolis, MN 55455 \\
\texttt{haeni001@umn.edu} \\
\And
Pravakar Roy \\
Department of Computer Science \& Engineering\\
University of Minnesota\\
Minneapolis, MN 55455 \\
\texttt{royxx268@umn.edu} \\
\AND
Volkan Isler \\
Department of Computer Science \& Engineering\\
University of Minnesota\\
Minneapolis, MN 55455 \\
\texttt{isler@umn.edu} \\
}

\usepackage{soul}
\usepackage[dvipsnames]{xcolor}
\usepackage{cite}

\begin{document}

\maketitle

\begin{abstract}
We present a modular end-to-end system for yield estimation in apple orchards. Our goal is to identify fruit detection and counting methods with the best performance for this task. We propose a novel semantic segmentation based approach for fruit detection and counting and perform extensive comparative analysis against other state-of-the-art techniques. This is the first work comparing multiple fruit detection and counting methods head-to-head on the same datasets. Fruit detection results indicate that the semi-supervised method, based on Gaussian Mixture Models, outperforms the deep learning based methods in the majority of the datasets. For fruit counting though, the deep learning based approach performs better for all of the datasets. Combining these two methods, we achieve yield estimation accuracies ranging from $95.56\%-97.83\%$.
\end{abstract}
\section{Introduction}

\begin{figure*}[ht!]
	\centering
	\includegraphics[width=\textwidth]{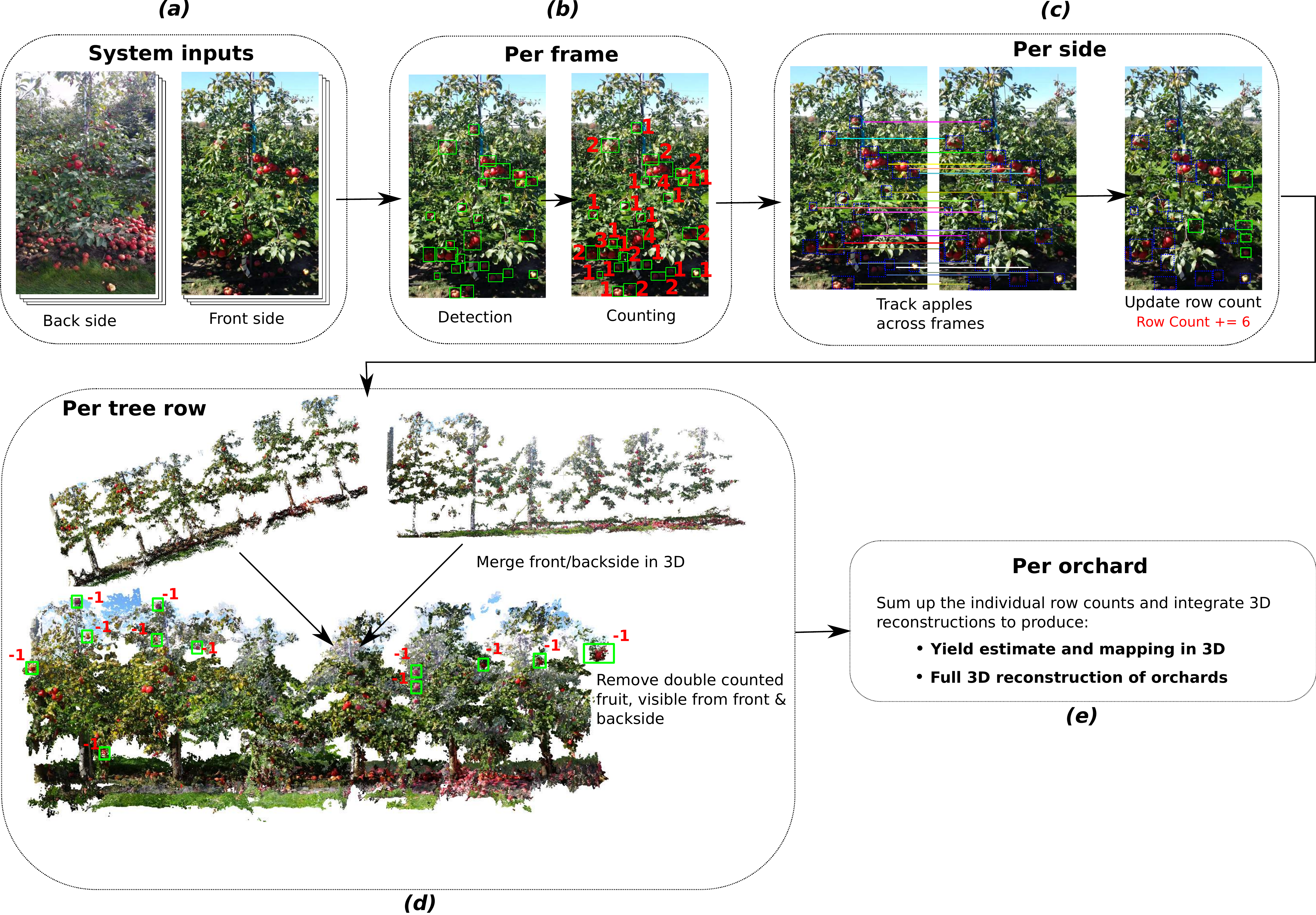}
	\caption{Overview of the yield estimation process. (a) Given two image sequences from the same portion of an orchard row, one from the front and one from the back. (b) Fruits are detected and counted in each frame. (c) Fruits are tracked across the image sequence to avoid double counting. As outputs, we have an image sequence per side, together with fruit locations and counts. (e) We reconstruct each image sequence in 3D and merge the two reconstructions into a single 3D model of the tree row. For yield estimation, we can now remove fruits visible from both sides of the tree row. (f) The pipeline produces yield estimates together with a 3D reconstruction and fruit mapping.}
	\label{fig:concept}
\end{figure*}

Precision agriculture techniques have ushered in an unprecedented era of automation in farming and food production. There exist now commercial solutions for automating numerous farming tasks such as monitoring crop fields (\cite{gamaya_gamaya_2019}), seeding (\cite{tanimura_plant_2019}), harvesting (\cite{cnh_cnh_2019}), and packaging. However, the application domain of precision agriculture has been primarily limited to commodity crops such as wheat, rice, maize, etc. Developing solutions for specialty crops (such as fruits or vegetables) has been challenging due to the complex geometry of orchards compared to row crops. As it is hard to generalize automation techniques across multiple specialty crops, researchers have focused on crop specific systems. As it is hard to generalize automation techniques across crops, researchers have focused on developing systems for specific crops. For apples, researchers have addressed problems such as diameter estimation~\cite{wang_automated_2013}, automated pruning~\cite{he_sensing_2018} or apple picking~\cite{laugier_autonomous_2008}. One of the precursors to these tasks is the accurate detection and localization of fruits on trees. In his work, we focus on establishing fruit detection, localization, counting and tracking methods for yield mapping.

Automated yield mapping is a precursor to numerous precision farming and phenotyping tasks. In the absence of an automated system, growers have to resort to manual data collection. Workers have to sample a few trees at random and extrapolate the counts to the whole orchard, which can lead to erroneous yield estimates. Consequently, significant research has been dedicated to automating the yield mapping process~\cite{bargoti_image_2017, wang_automated_2013, gongal_apple_2016, das_devices_2015, stein_image_2016, roy_registering_2018, roy_vision-based_2018}. In general, these yield mapping systems collect images from a single/both sides of a tree row, detect and count the fruits, track them across multiple frames and merge the fruit counts. Before our work in~\cite{roy_registering_2018}, none merged fruit counts from both sides of the tree row without the help of external navigational sensors. In contrast, we presented a method using semantic information to build a consistent model of the tree row and used the merged reconstruction to combine fruit counts from both sides. This method achieved yield estimation accuracies ranging from $91.98\% - 94.81\%$. An overview of this system is given in Figure~\ref{fig:concept}. In this paper, we use the same end-to-end system for yield mapping but improve several key components of the system, resulting in yield estimation accuracies of $95.56\%-97.83\%$. Additionally, we provide extensive experimental evaluations and conduct a comparative study with other state-of-the-art methods. Our main technical contributions are the following:

\begin{description}
	\item [Fruit detection] We demonstrate how a semantic segmentation network (U-net) can be used to identify apples and provide performance comparisons with other state-of-the-art methods. 
	
	\item[Fruit counting] We present an improved method over~\cite{hani_apple_2018} for estimating accurate fruit counts at the image level.
	
	\item [Comparative study of state-of-the-art fruit detection and counting methods] New apple detection and counting approaches proposed in the literature are tested on separate, specialized datasets, making head-to-head comparisons of the results impossible. To alleviate this problem, we evaluate the performance of three fruit detection and two counting methods on the same datasets. We hope that in the future our experiments together with the datasets can form a benchmarking platform.
	
	\item [Yield estimation] Using the system presented in our recent work~\cite{roy_registering_2018} we provide complete yield results.
\end{description} 

In Section~\ref{sec:relwork} we present related work on the topic of yield estimation, fruit detection, counting, and tracking. The problem formulation together with an overview of the end-to-end yield estimation process used for this paper is presented in Section~\ref{sec:formulation}. Section~\ref{sec:method} presents the technical details of the individual components. Experimental results, together with the strengths and weaknesses of each approach, are shown in Section~\ref{sec:experiments}. 
\section{Related Work}\label{sec:relwork}
Yield estimation for specialty crops is a challenging problem, which has received much attention over the last few years. The developed approaches range from complete processing pipelines to proposals for a single component, such as fruit detection, counting or tracking. We first discuss standalone yield estimation systems. Next, we take a closer look at approaches that address the fruit detection, counting, and tracking parts. 

\subsection{Systems for Yield Estimation}
Hand-designed algorithms used thresholding techniques together with color/shape features to detect fruits. Many of these systems relied on controlled environments for stable fruit detection. \cite{wang_automated_2013} proposed a system that operates at night with artificial flashlights. To control illumination conditions, \cite{gongal_apple_2016} developed an over-the-tree sensor system.
Fruits in these works were detected using static or dynamic thresholding~\cite{otsu_threshold_1979}. The resulting binary masks were identified as fruits using static features such as eccentricity of Circular Hough Transform~\cite{pedersen_circular_2007}. For tracking, the first method used a navigational sensor and triangulation, while the second used the 3D information of a time-of-flight camera.

Currently, we see a drastic change in algorithm design in the domain of precision agriculture through the adoption of machine learning techniques. \cite{das_devices_2015} used a Support Vector Machine (SVM) to classify each image pixel into fruit/background. The detected fruits were tracked across frames using optical flow. To compensate for miscounted fruits, they counted fruits manually on $26$ trees and fit a linear model to the data, to adjust the estimated counts. \noindent\cite{roy_vision-based_2018} used a Gaussian Mixture Model (GMM), both for fruit detection and counting. Using the tracking method in ~\cite{roy_registering_2018}, they reported total yield estimates achieving accuracies ranging from $91.98\% - 94.81\%$. 

More recently, deep learning has transformed the field of computer vision and precision agriculture. An early adopter of neural networks, \cite{stein_image_2016} proposed a multi-sensor framework to count mangoes. The fruits were detected using a Faster R-CNN (FRCNN)~\cite{ren_faster_2015} network. The position of the mango fruit was estimated using epipolar geometry and GPS information. Additionally, each fruit was assigned to a single tree, using LIDAR generated tree masks. \cite{bargoti_image_2017} used Multilayer Perceptrons and a Convolutional Neural Network (CNN) to segment images into apple/background pixels. The segmented fruits were counted using a combination of Circular Hough Transform and Watershed Transform~\cite{beucher_watershed_1992}. They sampled images manually at roughly 0.5m to minimize image overlap and avoid double counting. The total yield was estimated by summing up the counts of the individual images. \cite{liu_robust_2018} used a Fully Convolutional Network (FCN)~\cite{long_fully_2015} to segment images into fruit/background pixels. The individual fruits were tracked across frames using a Kanade-Lucas-Tomasi (KLT) tracker~\cite{lucas_iterative_1981} together with the Hungarian algorithm~\cite{kuhn_hungarian_1955}. A Structure from Motion (SfM) algorithm was used to compute relative 3D fruit locations and size estimates.

Most of the systems presented so far, partially follow our outline shown in Figure~\ref{fig:concept} and contain a similar set of components. However, most of them do not build a coherent geometric model from both sides of the row and instead try to find a consistent relationship between single side fruit counts and actual yield. Such systems are bound to overestimate/underestimate fruit counts, specifically in environments where the trees are not well pruned. 

\subsection{Fruit Detection}
The first step in a yield mapping pipeline is the accurate detection of fruits.
Early methods mostly relied on static color thresholds for detection. The limitations of these methods were often concealed by adding additional sensors, such as thermal- or Near Infrared (NIR) cameras. \cite{gongal_sensors_2015} offers a comprehensive overview of these early detection methods. More recently,~\cite{hung_feature_2015} used a Conditional Random Field (CRF) to segment the image into 5 classes (fruits, trunk, leaves, ground, and sky). Their model achieved F1 detection scores of $0.81$. In 2015, Faster R-CNN~\cite{ren_faster_2015} became the de-facto state-of-the-art object detection method.  Two recent papers used this network for fruit and vegetable detection. \cite{bargoti_deep_2017} used an FRCNN to detect and count almonds, mangoes, and apples, achieving $F_1$-scores of $>0.9$. \cite{sa_deepfruits:_2016} used a similar approach to detect green peppers. They merged RGB and Near-Infrared (NIR) data and achieved $F_1$-scores of $0.84$.

In contrast to these methods, we treat fruit detection as a pixel-wise classification problem and train a semantic segmentation network (U-net) for solving it. We perform a rigorous performance comparison with two state-of-the-art fruit detection methods - a classical semi-supervised approach using GMM~\cite{roy_vision-based_2018} and an object detection approach based on Faster R-CNN~\cite{bargoti_deep_2017}. 

\subsection{Fruit Counting}
We have mentioned many different approaches for fruit counting already. These approaches often used classical (Circular Hough Transform (CHT)) and object-detection based counting. Using state-of-the-art object detection networks does not require additional counting algorithms. Instead, the network outputs are summed up to receive counts~\cite{bargoti_deep_2017, sa_deepfruits:_2016, stein_image_2016}. 
Both of these popular counting methods have drawbacks. The main drawbacks of CHT are its reliance on accurate segmentation of the image, inability to handle occlusions and the need to fine tune parameters across datasets. Object detection networks use Non-Maximum Suppression to reject overlapping instances, and often true positives are filtered out. \cite{bargoti_deep_2017} found that up to $4\%$ of their error was due to this filtering process. 

Alternatively, \cite{chen_counting_2017}, used a combination of two networks for fruit counting. First, an FCN creates feature maps of possible targets in the image. Second, a CNN counts these targets using a regression head.
\cite{maryam_rahnemoonfar_deep_2017} used a CNN for direct counting of red fruits from simulated data. They used synthetic data to train a network on 728 classes to count red tomatoes. Different to these approaches, \cite{roy_vision-based_2017} modeled fruit clusters using Gaussian mixture models and presented a method for fruit counting by finding the optimal number of components in the model. This approach is entirely unsupervised, demonstrated qualitative results and achieved an overall counting accuracy of $94.4\%$. \cite{roy_active_2017} presented a method for counting fruits using a robotic manipulator for active perception.

In contrast to these methods, we formulate fruit counting as a multi-class classification problem. We train a CNN on image patches that represent apple clusters which are then counted by the network in a classification manner with high accuracy. This method is an improvement of our previous work in~\cite{hani_apple_2018}. In this work, we modified the old network to improve accuracy and added extensive experimental validation. 

\subsection{Fruit Tracking}
After the fruits are detected and counted, they need to be tracked across frames to avoid double counting. \cite{wang_automated_2013} used stereo block matching and GPS information to register the fruit locations globally. Similarly, \cite{stein_image_2016} used GPS position together with epipolar geometry to accurately track fruits. \cite{hung_feature_2015} avoided the tracking problem altogether, by choosing a sampling frequency so that images would not overlap at all. While this approach is easy to implement, our previous work in \cite{roy_vision-based_2018} found that using multiple views greatly increases the robustness of fruit detection and counting. \cite{das_devices_2015} used a multi-sensor platform and optical flow to avoid double counting. In our previous work in~\cite{roy_surveying_2016}, we presented a method for registering apples based on affine tracking~\cite{baker_lucas-kanade_2004} and incremental Structure from Motion (SfM)~\cite{sinha_multi-stage_2012}.

Since fruits are often visible from both sides of a tree row, it is essential to merge the fruit counts from both sides. In our previous work in~\cite{roy_registering_2018}, we addressed this problem. We used semantic information to merge the front and backside of a tree row, and we demonstrated the necessity of this step by analyzing the fruit counts with and without this alignment.
\section{Problem Formulation and Overview of the Entire System}\label{sec:formulation}
\textit{``Given two captured image series of the same portion of a tree row from a monocular camera (i.e. cell phone, GoPro etc)- one from the front side of the row and one from the back side- we want to estimate the total fruit counts and locations for the captured portion of the row''}. 

An end to end solution to this problem involves solving multiple subproblems such as fruit detection, counting, tracking fruits across multiple views and merging fruit counts from both sides of a row. Many approaches strive to solve this problem - some rely on specialized sensors and hardwares~\cite{das_devices_2015,gongal_apple_2016}, some correlate fruit counts from a single side to ground truth \cite{stein_image_2016}. In contrast to these works, we follow the approach proposed by~\cite{roy_registering_2018, dong_semantic_2018}. Their system detects and counts apples in individual images, and the fruits are tracked across an image sequence using estimated camera motion. Each side of a row is reconstructed independently and merged into a coherent 3D model. With the help of this model, the system eliminates duplicate fruit counts owing to fruits visible from both sides. This is the first approach that can merge fruit counts from both sides of a tree row without relying on any specialized hardware. 

\subsection{Per Frame Fruit Detection and Counting}
The per frame fruit detection and counting component takes an individual frame as input and outputs a set of detected fruit clusters and corresponding fruit counts. In this work, we analyze the performance of multiple detection and counting algorithms to find the best-suited method for yield estimation. 
For detection we propose a novel approach using an image segmentation network (U-Net~\cite{ronneberger_u-net:_2015}). We compare this network to an object detection network~\cite{bargoti_deep_2017} and a color-based clustering technique using Gaussian Mixture Models (GMM)~\cite{roy_vision-based_2017}. We perform an extensive experimental evaluation of all three methods. These results are presented in Section~\ref{subsec:detection_result}.

For counting of clustered fruits, we evaluate two approaches. First, we present an improved deep learning based counting approach, for which we presented initial results in~\cite{hani_apple_2018}. Second, we briefly review a classical method of counting by using GMM and image segmentations~\cite{roy_vision-based_2017}. We perform a comparative analysis of the performance of these methods in Section~\ref{subsec:count_result}.

\subsection{Tracking Fruits and Merging Fruit Counts Across Multiple Views}
Tracking fruits is an essential task to avoid over counting. The tracking component takes the entire sequence of images from a single side of a tree row and the per-frame fruit detections and counts as input. We use the previously proposed method in~\cite{roy_registering_2018, dong_semantic_2018} for fruit tracking. For completion, we review this component in Section~\ref{subsec:tracking}.

\subsection{Merging Fruit Counts from Both Sides and Yield Estimation}
This component takes the single side reconstructions, and multi-view fruit counts as inputs. It merges the input reconstructions from both sides using semantic information~\cite{roy_registering_2018,dong_semantic_2018}, eliminates double counting owing to fruits visible from both sides of the tree and outputs the total fruit count for the captured portion of the row. Again, for completion, this component is briefly reviewed in Section~\ref{subsec:tracking}.
\section{Technical Approach}\label{sec:method}
In this section, we provide details for each of the components of the end-to-end yield estimation system discussed in the previous section. 

\subsection{Fruit Detection}\label{sec:detection}
At present, deep-learning-based approaches are dominating the field of image segmentation and object detection~(\cite{badrinarayanan_segnet:_2017, ronneberger_u-net:_2015}). They have been effective in fruit and crop segmentation as well~(\cite{chen_counting_2017, hani_apple_2018}). The overwhelming success of these techniques is often attributed to the huge amount of training data from which the networks learn features that ideally generalize across environments. Obtaining such data in cluttered environments such as orchard settings though is hard and cumbersome. Labeling a $1920\times1080$ image can take up to $15$ minutes depending on the number of apples in it. Therefore, it is important to quantify the improvement in detection and counting performance by deep network-based models compared to simpler, classical methods. 

Toward this goal, we test three different methods for fruit detection. First, we present a novel approach based on a pixel-wise segmentation network, U-Net~\cite{ronneberger_u-net:_2015}. We provide details for implementation and rationale for design choices. Second, we revisit an object detection network presented by~\cite{bargoti_deep_2017}, which used a Faster R-CNN~\cite{ren_faster_2015} for object detection. For this, we provide reimplementation details with several improvements, such as changing the backbone to a deeper network and the inclusion of a focal loss term~\cite{lin_focal_2017}. Finally, we review a semi-supervised color-based clustering technique using Gaussian Mixture Models (GMM) and Expectation Maximization (EM)~\cite{moon_expectation-maximization_1996} presented in \cite{roy_vision-based_2018}. We perform a thorough side-by-side evaluation of these methods in Section~\ref{subsec:detection_result}.

\subsection{Fruit Detection by Semantic Segmentation}\label{subsec:unetdetection}
We use a semantic segmentation network to assign a class (apple/background) to each pixel, while minimizing the classification error. The literature contains different networks for dense semantic segmentation~\cite{long_fully_2015,badrinarayanan_segnet:_2017}. However, to be effective for fruit detection, a solution must have specific capabilities:

\begin{itemize}
	\item As data annotation for fruit trees is difficult, the network architecture must be able to leverage small amounts of training data and use the available data efficiently.
	
	\item The design must be able to deal with small objects and occlusions. From a typical imaging distance of $2-3$ meters, apples often occupy $5 - 50$ pixels in a $1920 \times 1080$ image.
	
	\item The network must be able to handle class imbalance since the ratio of fruit to background pixels is roughly $1:20$
\end{itemize}

\begin{figure*}[!hbpt]
	\centering
	\def\svgwidth{\textwidth}
	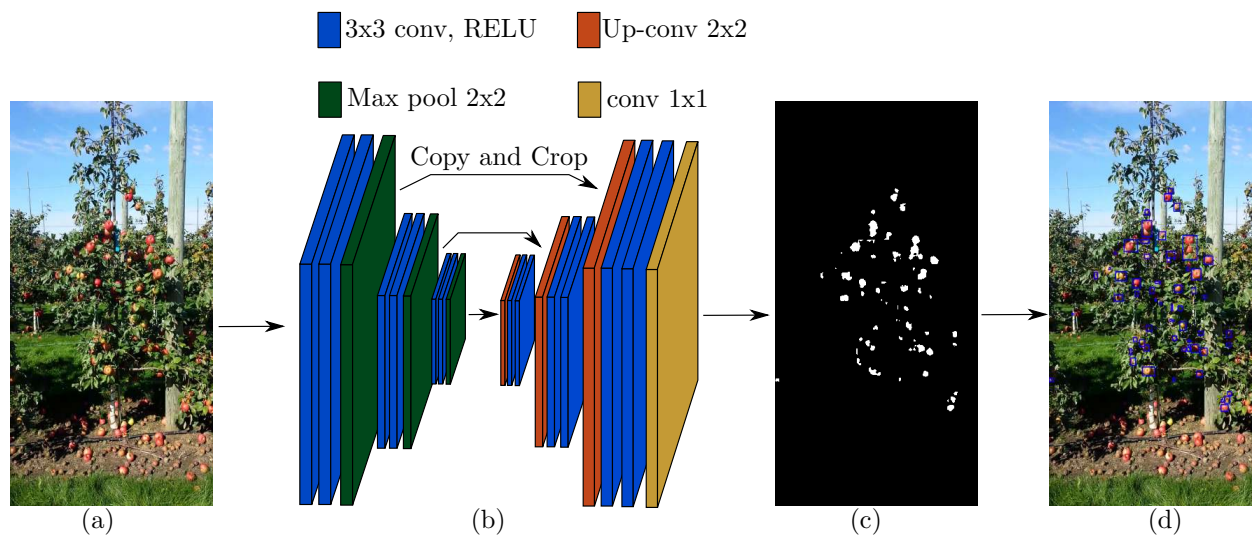\label{fig:unet}
	\caption{Schematic drawing of U-Net detection work flow. (a) Input image; (b) Schematic view of the down/up-sampling in U-Net~\cite{ronneberger_u-net:_2015}; (c) The per pixel segmentation of U-Net; (d) Detections after connected component analysis.}
	\label{fig:unet_pipeline}
\end{figure*}

A Convolutional Neural Network (CNN) that achieves high precision and recall for small objects with a small amount of data is U-Net~\cite{ronneberger_u-net:_2015}. U-Net contains a contracting and expanding path, where the contracting path is a typical convolutional network. During the contraction, the spatial information is reduced while feature information is increased. The expansive pathway combines the feature and spatial information through a sequence of up-convolutions and concatenations (skip-connection layers) with high-resolution features from the contracting path. To deal with the problem of class imbalance, we used weighted categorical cross-entropy as our loss function. This loss function allows setting weights that depend on the type of misclassification. For each pair of class labels, $X$ and $Y$ one may specify how to penalize label a misclassification of label $X$ when the actual label is $Y$. In our case, the class weights are chosen to be inversely proportional to their frequency in the training data. Additionally, the class weights are normalized. For a schematic overview of our proposed method using this U-Net architecture see Figure~\ref{fig:unet_pipeline}.

\textbf{Implementation Details:} All the networks in this work were trained on machines of the Minnesota Supercomputing Institute (MSI) at the University of Minnesota. We used a single node, of which each contains $2$ NVIDIA Tesla K20X GPUs, each with $12$ GB of memory. 
The network was implemented in Tensorflow~\cite{abadi_tensorflow:_2015}, using the Keras~\cite{chollet_keras_2015} Application Programming Interface (API). We used a VGG-16~\cite{simonyan_very_2015} network as the backbone in the contracting path. The classification layers were replaced with up-convolutions and concatenation layers for the expansive path. We initialized the contractive path with pre-trained weights of the ImageNet dataset~\cite{russakovsky_imagenet_2015}. We used images of dimensions $224 \times 224$ and trained with a batch size of $32$ images per GPU for up to $50$ epochs. We used Ada-Delta as the optimizer and did not apply any image augmentations.

\subsubsection{Fruit Detection by Object Detection Network}
Object detection networks simultaneously localize and classify an unknown number of objects in an image. In the literature, two meta-architectures have emerged: Single Shot Multibox Detectors (SSD)~\cite{leibe_ssd:_2016} and two-stage detectors, such as Faster R-CNN~\cite{ren_faster_2015}. Single Shot Detectors use a single feed-forward network to predict object class scores together with bounding box proposals. Two-stage detectors consist of a Region Proposal Network (RPN) and additional network heads. The RPN proposes Regions of Interest (RoI). The second stage computes a classification score together with class-specific bounding box regression for each of these regions. While SSD based models are faster on average than two-stage models, this comes at the cost of detection accuracy~\cite{huang_speed/accuracy_2017}. This effect is even more pronounced when the objects in question are small. For this reason, mainly two-stage networks have been adopted for the task of fruit detection~(\cite{sa_deepfruits:_2016, stein_image_2016, bargoti_deep_2017}). 

\begin{figure*}[!hbpt]
	\centering
	\def\svgwidth{\textwidth}
	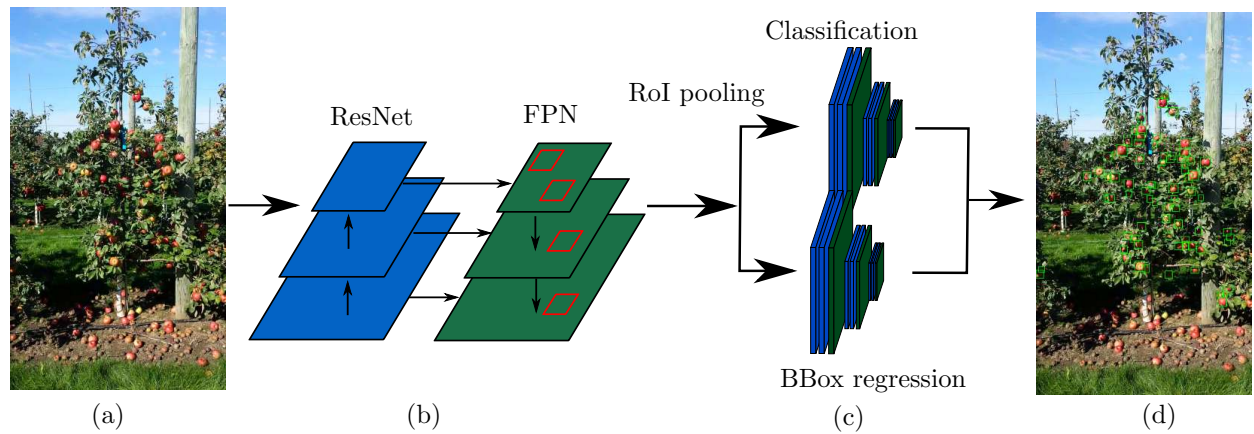\label{fig:frcnn}
	\caption{Schematic drawing of Faster R-CNN with FPN head and focal loss. (a) Input image; (b) Region proposal network, using ResNet50 as backbone~\cite{lin_feature_2017}; (c) Network heads for classification and regression on object detection location; (d) Detections.}
	\label{fig:frcnnpipeline}
\end{figure*}

For this study, we reimplemented the network proposed by~\cite{bargoti_deep_2017}. We did not consider~\cite{sa_deepfruits:_2016} since they used a combination of RGB and NIR data for prediction. FRCNN has been shown to perform poorly when detecting objects smaller than $32 \times 32$ pixels. To counteract this problem, we added a Feature Pyramid Network (FPN) with lateral connections~\cite{lin_feature_2017} to the detection stage of the network. Another modification that we included in the network was a focal loss term, as proposed by~\cite{lin_focal_2017}. This focal loss is added to the standard cross-entropy term and reduces the loss for easily classifiable examples. For a schematic overview of the network see Figure~\ref{fig:frcnnpipeline}.

\textbf{Implementation Details:} The network was trained on a single node, which contains two NVIDIA Tesla K20X GPUs. We used a Tensorflow implementation of Faster R-CNN, which can be found online\footnote{\url{https://github.com/fizyr/keras-retinanet}}. We followed the parameters used in~\cite{bargoti_deep_2017, ren_faster_2015}. The  anchor scales were modified to: $\lbrace 32, 64, 128, 256, 512 \rbrace$. \cite{bargoti_deep_2017} used VGG~\cite{simonyan_very_2015} as backbone architecture. In this work, we use a ResNet50~\cite{he_deep_2016}. We trained the network for a maximum of $100$ epochs, each with $10000$ iterations. Images were resized so that the larger size is $500$ pixels. We only used left/right flipping as data augmentations. 

\subsubsection{Fruit Detection by Semi-supervised Clustering} \label{subsec:gmmdetection}
It is important to quantify the improvement in detection and counting performance by deep network-based models compared to simpler, classical methods. For this purpose, we previously described methods against~\cite{roy_vision-based_2018}, who presented a simple method for apple detection, based on semi-supervised clustering with Gaussian Mixture Models (GMM). Here, we briefly revisit the method for completion.

\begin{figure*}[!hbpt]
	\centering
	\subfloat[][Input image]{\includegraphics[width=0.23\textwidth]{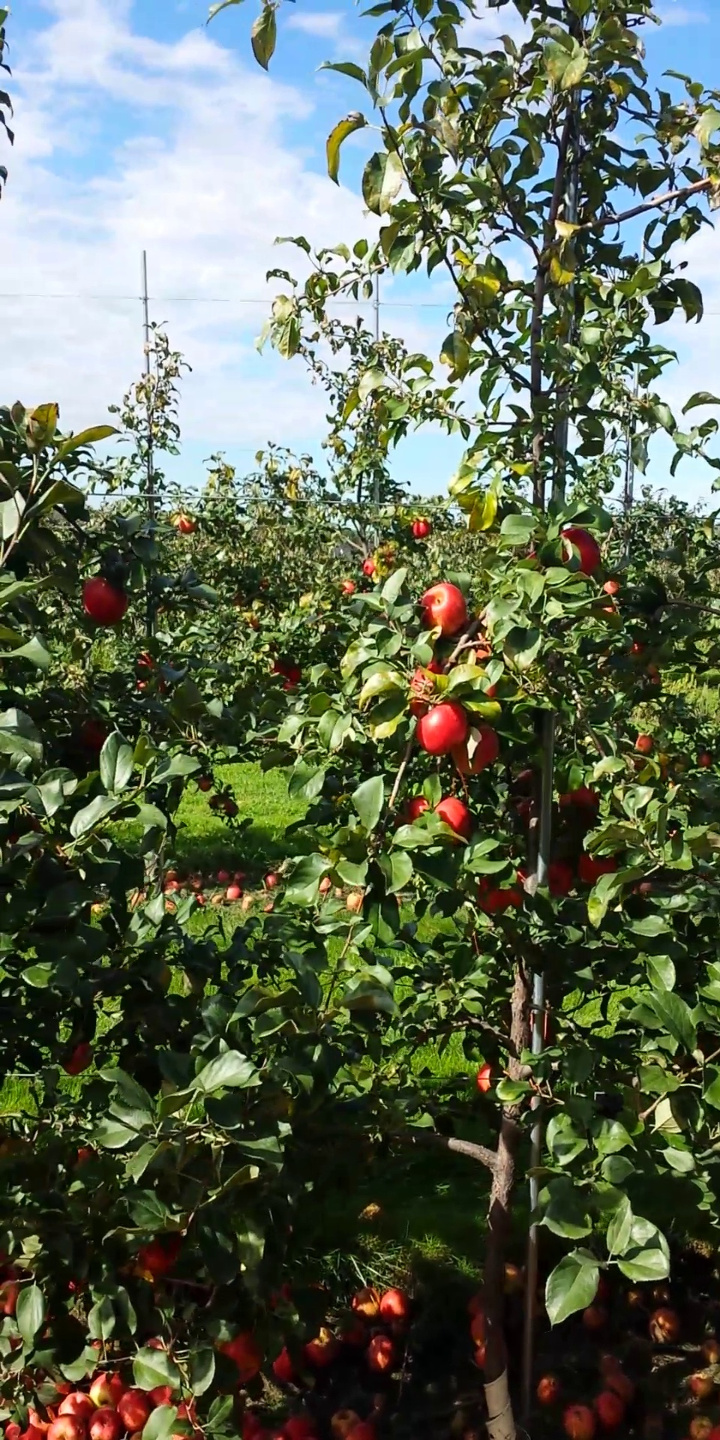}\label{fig:gmm1}}
	\quad
	\subfloat[][Superpixels]{\includegraphics[width=0.23\textwidth]{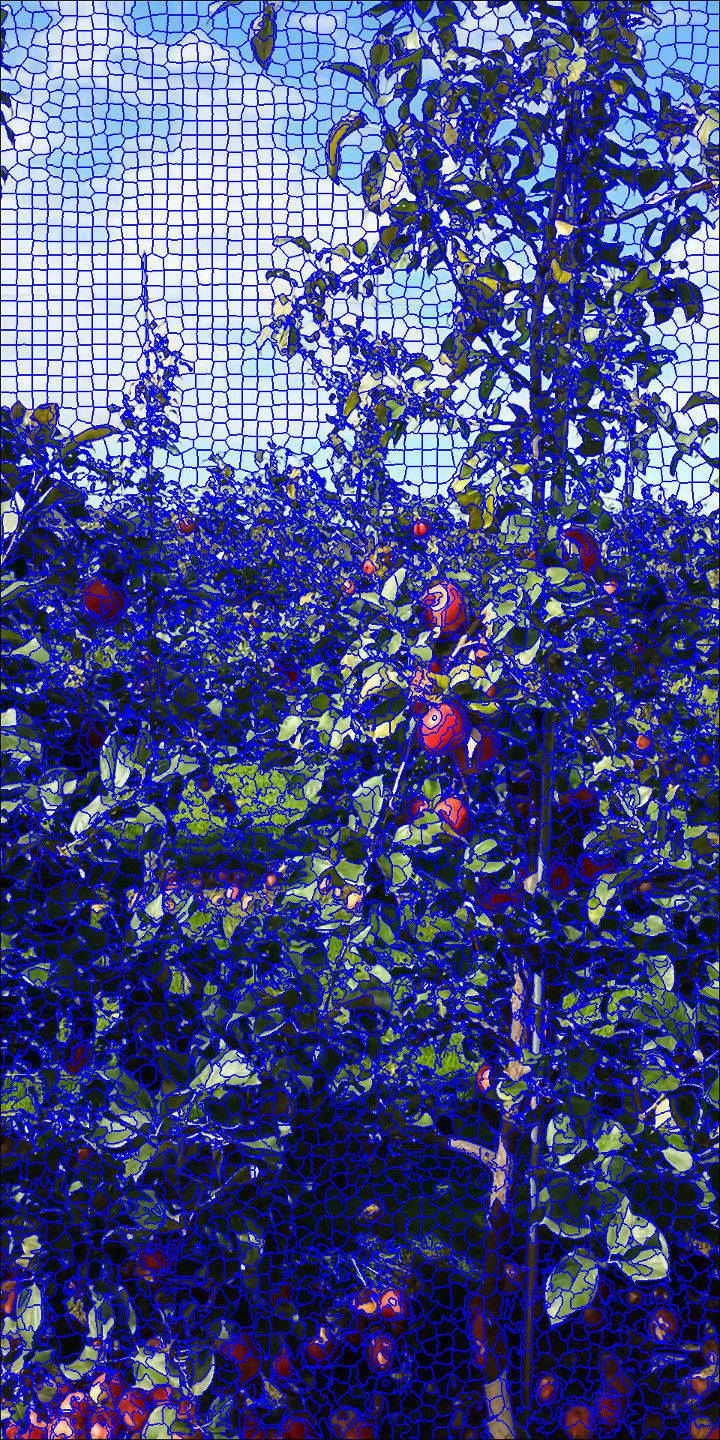}\label{fig:gmm2}}
	\quad
	\subfloat[][Segmented image]{\includegraphics[width=0.23\textwidth]{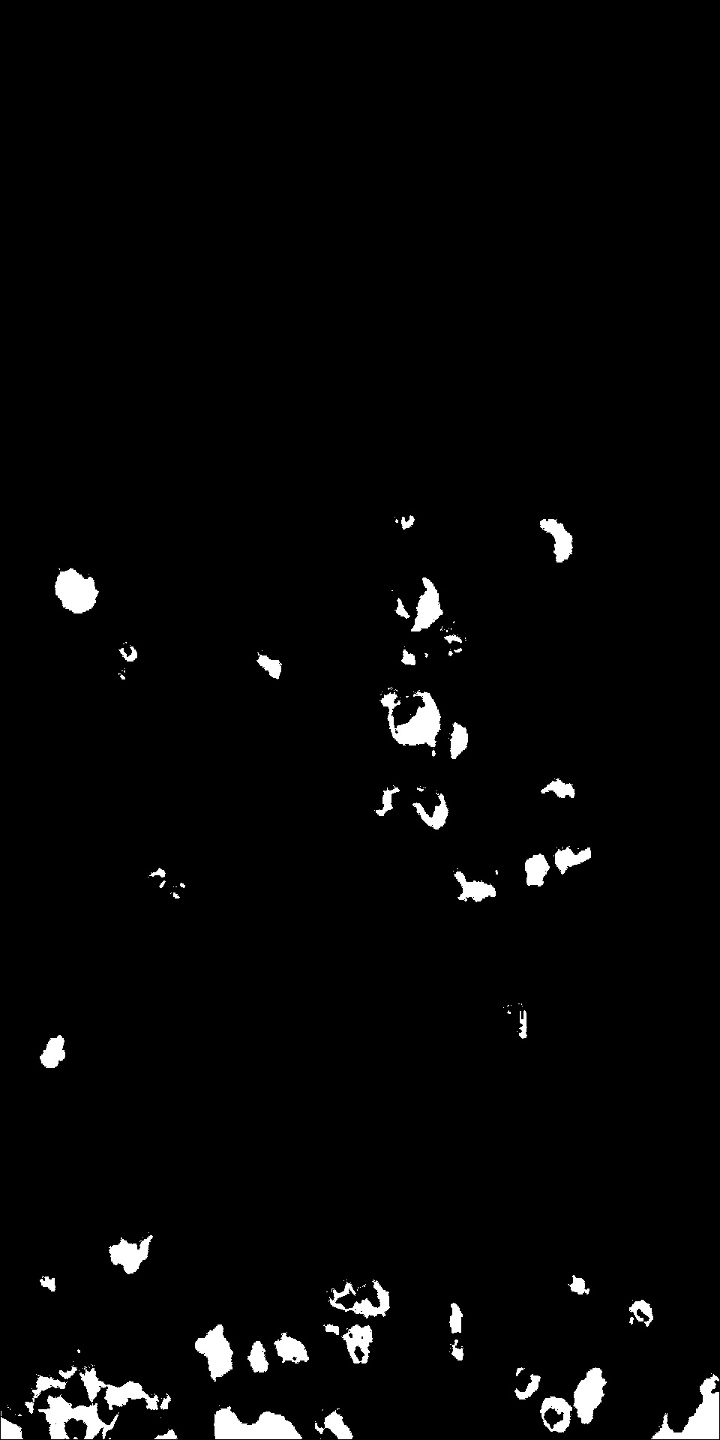}\label{fig:gmm3}}
	\quad
	\subfloat[][Detections]{\includegraphics[width=0.23\textwidth]{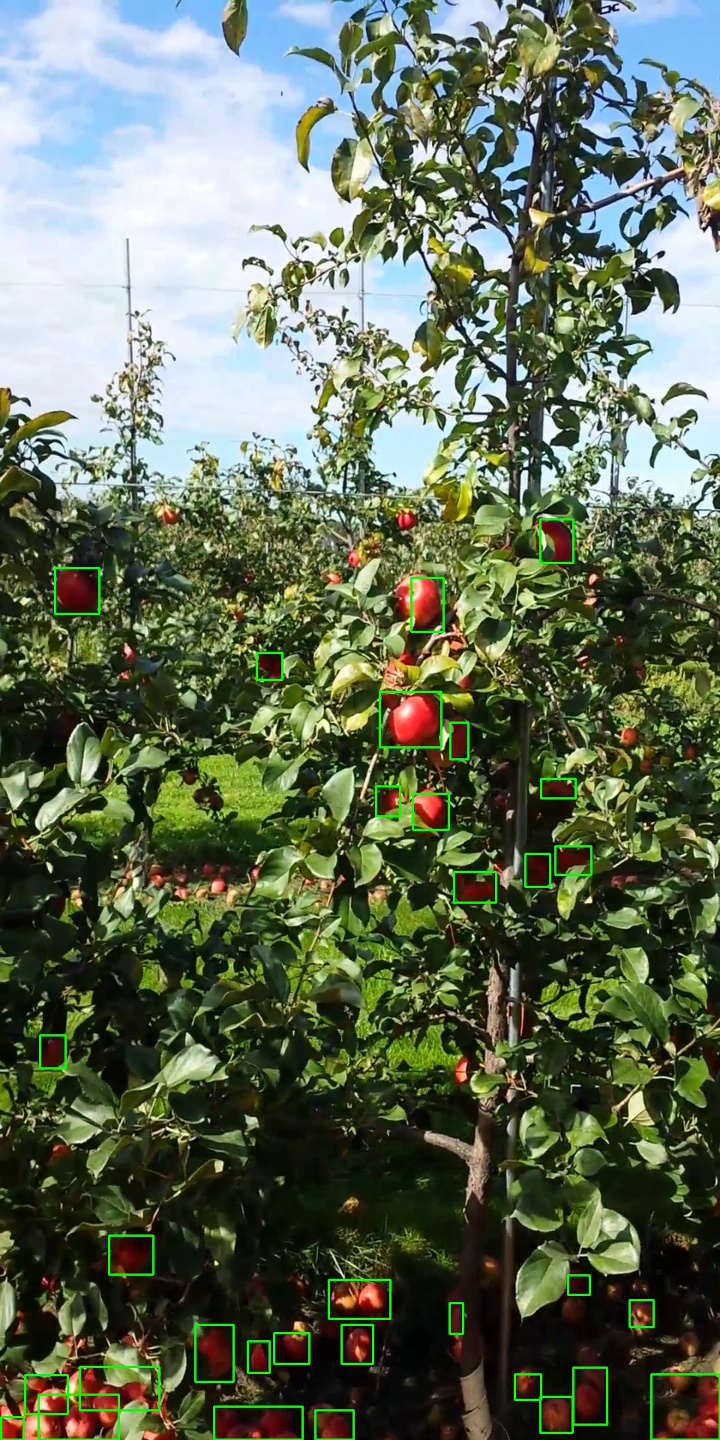}\label{fig:gmm4}}
	\caption{Semi-supervised detection pipeline from~\cite{roy_vision-based_2018}.}
	\label{fig:seg_pipeline}
\end{figure*}

The method over-segments the input image into SLIC super-pixels~\cite{achanta_slic_2010}, using the LAB colorspace. Each super-pixel is represented by the mean LAB color of the pixels within the super-pixel. The resulting super-pixels are clustered into approximately $25$ color classes. Finally, each super-pixel is classified into apple or background, based on KL divergence~\cite{goldberger_efficient_2003} from hand-labeled classes. These classes are obtained by user-supervision. The method provides a user interface, where the user is asked to provide supervision for a few frames, by identifying the classes belonging to apples. The user-supervision allows this method to account for different lighting conditions and the color of the particular apple variety. The steps of this method can be found in Figure~\ref{fig:seg_pipeline}.

The main advantage of this method is the simplicity of annotation. Typically, $10-20$ clicks (user-supervision) are enough to create a classification model to detect all the fruits in a video. However, this method does not work if the fruits are not distinguishable by color from the background. The training model for the GMM is obtained in a user-supervised or semi-supervised fashion. User-supervised means that the first 5 frames of a given dataset were used to train the model. The semi-supervised version of the model was created on a single training dataset, different from the test sets. In section~\ref{subsec:detection_result}, we evaluate both of these techniques against the deep learning counterparts.

\subsection{Fruit Counting}\label{sec:counting}
Once the fruits are detected, we aim to count them in a per frame manner. Challenges that make this task difficult are: (1) Apples often grow in arbitrarily large clusters. (2) Apples are often occluded by branches, leaves or other fruits. As we discussed in Section~\ref{sec:relwork}, existing methods for fruit counting are dominated by Circular Hough Transformation (CHT)~\cite{pedersen_circular_2007}. CHT requires extensive parameter tuning and fails to handle occlusions. These issues led to the development of more sophisticated methods for fruit counting.
In this section, we discuss two such approaches. First, we present an approach where we formulate the fruit counting problem as a multi-class classification task and solve it using a CNN. A preliminary version of this work was presented in \cite{hani_apple_2018}. In this version, we change the backbone network to a deeper Resnet50 and add extensive experimental evaluation. Second, we review a classical approach using GMM and Expectation Maximization (EM)~\cite{roy_vision-based_2017, roy_vision-based_2018}. Both of these approaches work with either the semi-supervised or the U-Net detection methods as inputs. With the Faster R-CNN method, fruit counts are equivalent to summing up the individual detections.


\subsubsection{Counting using Convolutional Neural Network}
\label{sec:countingCNN}
We approach the problem of accurately estimating clustered apple counts by making the following observations: (1) Apples are sparsely distributed over the whole image; (2) they are often clustered together; (3) and cluster sizes are unevenly distributed. We define the accurate counting of clustered apples as a classification problem with a finite number of classes representing the apple counts per Region of Interest (RoI). This method takes the detected ROI's, that are likely to contain apples, as inputs. The network can receive input from a variety of detection algorithms. For example, any of the methods described in Section~\ref{sec:detection} would produce acceptable image patches. We constrain the maximum size of the apple clusters to six apples. Previous work~\cite{roy_vision-based_2018, hani_apple_2018} has empirically established a cluster size of six apples as a reasonable upper bound. We, therefore, define $7$ classes, representing the apple counts per image patch, including zero. 

\begin{figure*}[!hbpt]
	\centering
	\def\svgwidth{0.95\textwidth}
	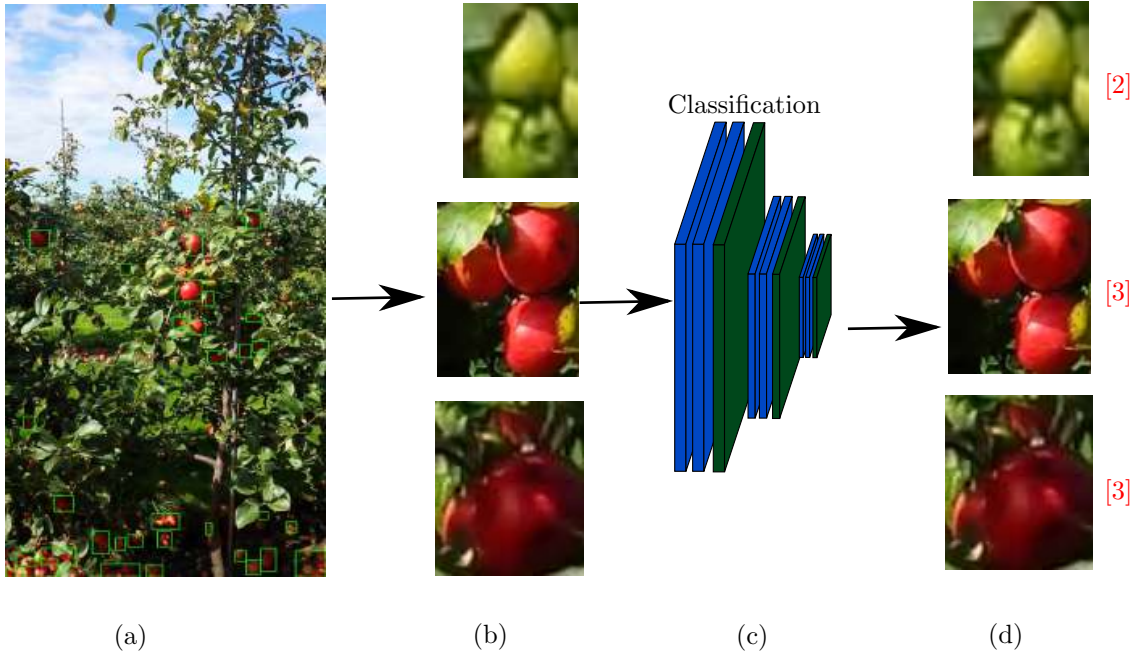\label{fig:cnn}
	\caption{Schematic drawing of counting using classification network. (a) Input image with detected regions of interest; (b) Extracted regions; (c) ResNet50 for classification of image patches; (d) Patches are classified into 7 counts.}
	\label{fig:cnnpipeline}
\end{figure*}

In ~\cite{hani_apple_2018} we presented a preliminary version of this method using an AlexNet~\cite{krizhevsky_imagenet_2017} architecture. In this work, we used a ResNet50~\cite{he_deep_2016} architecture. Additionally, we tested Google's Inception Resnet v2~\cite{szegedy_inception-v4_2017}. While this network contains $\sim56$ million parameters (compared to the $\sim26$ million of a ResNet 50), the network only performed $0.4\%$ better on average over all the validation sets. This improvement does not justify the size increase of $215$MB compared to the $99$MB of the ResNet50. Additionally, the Inception ResNet v2 took longer to train and was slower during inference.
We removed the fully connected and the prediction layers, before adding new ones for retraining. The image input layer was fixed to receive images of size $224\times224$ pixels and batch size was $64$. Images were loaded so that the larger size was equal to $224$ pixels. We kept the image's aspect ratio and padded the image with zeros if necessary. The weights of the network were initialized with pre-trained ones from ImageNet~\cite{russakovsky_imagenet_2015}. A schematic overview of the proposed counting approach can be seen in Figure~\ref{fig:cnnpipeline}.

\textbf{Implementation Details: }The counting network was trained on a single MSI node, which contains two NVIDIA Tesla K20X GPUs. The network was implemented in Tensorflow~\cite{abadi_tensorflow:_2015}, using Keras~\cite{chollet_keras_2015}. For optimization, we used the Adam optimizer with an initial learning rate of $0.001$, beta\_1 of $0.9$ and beta\_2 of $0.999$. We only used left/right flipping as image augmentations and trained the network for up to $50$ epochs. 

\subsubsection{Counting using Gaussian Mixture Model}
We compare the neural network counting method to one previously proposed in~\cite{roy_vision-based_2017}. We describe this method for completion. This method takes the segmented apple cluster images as input and outputs fruit counts and locations. It exploits the idea of the images being generated from a two-dimensional world model (probability distribution). Each apple in the input image is modeled by a Gaussian probability distribution function (pdf), and apple clusters are modeled as a mixture of Gaussians. The fruit counts are obtained from the world model configuration that is most likely to generate the image. If the correct number of apples are known a priori, we can find the most likely world model (GMM) using the EM algorithm~\cite{moon_expectation-maximization_1996} in this fashion. 
This method has the added advantage that it can compute locations of individual fruits in an input image. With the assistance of a depth camera, it has the potential to estimate fruit sizes. On the downside, it requires accurate segmentation of fruits and cannot reject false positives.

\subsection{Multi-view Fruit Tracking, Counting and Yield Mapping}\label{subsec:tracking} 
After the detection and counting stages, we have per frame fruit counts and their locations. To provide yield estimates, we need to integrate these counts over multiple views throughout the dataset. In addition to tracking the fruits from a single side, we need to register them from both sides of the tree row (some fruits are visible from both sides) to avoid double counting. In cluttered environments, such as apple orchards, the appearance of any given cluster changes drastically between views. It is possible that in some frames, the apples are partially occluded or not visible at all. By tracking clusters across frames and merging the individual predictions, we increase the robustness of our counting approach. 

\begin{figure*}[!hbpt]
	\centering
	\def\svgwidth{\textwidth}
	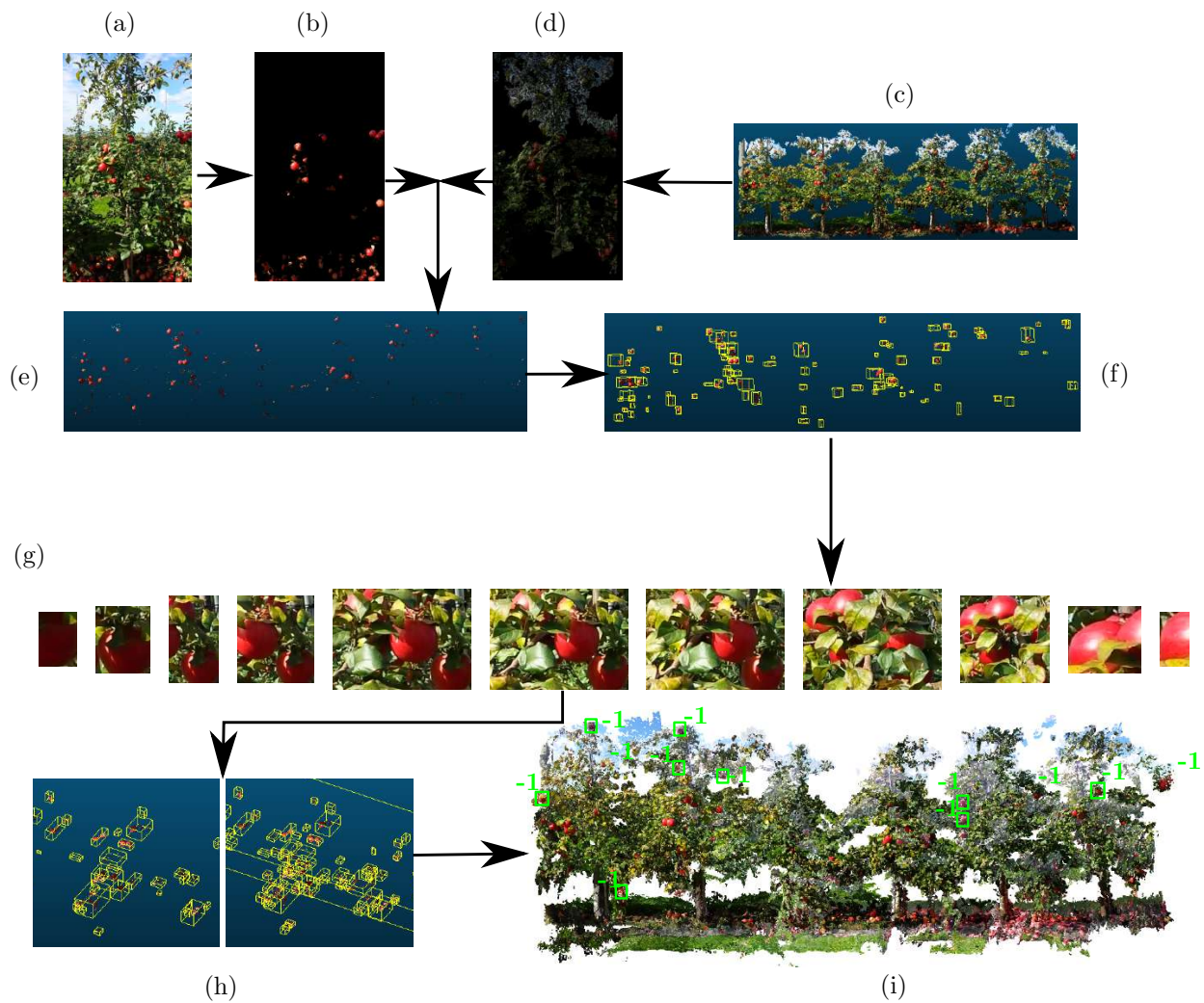\label{fig:track}
	\caption{Schematic overview of the fruit tracking and yield estimation pipeline. (a) Input image; (b) Segmentation of the fruits; (c) 3D Reconstruction of a single side of the tree row; (d) Reprojection of fruits on 2D images; (e) Segmented pointcloud; (f) 3D connected components in 3D; (g) Tracked fruit clusters reprojected across multiple frames; (h) Fruit clusters visible from both sides of a tree row; (i) Elimination of double counting for fruits visible from both sides.}
	\label{fig:trackpipeline}
\end{figure*}

The steps of the tracking algorithm are visualized in Figure~\ref{fig:trackpipeline}. These steps were developed as part of an extensive research effort, and we only mention the individual steps in this paper for completeness. To re-implement the multi-view tracking approach, we refer the interested reader to~\cite{roy_registering_2018, dong_semantic_2018}. Given an input image (Figure~\ref{fig:trackpipeline}(a)) we use the detection step to obtain a segmentation mask, containing only the fruits (Figure~\ref{fig:trackpipeline}(b)). Afterward, we use Structure from Motion (SfM) to obtain a semi-dense 3D reconstruction of the trees (Figure~\ref{fig:trackpipeline}(c)). However, the resulting point cloud does not have a one-to-one correspondence with the pixels in the images. To establish this correspondence, we project the reconstructed point cloud to all camera frames (Figure~\ref{fig:trackpipeline}(d)). We compute the intersection of the reprojected image and the binary mask to identify the 3D points belonging to the detected apples (Figure~\ref{fig:trackpipeline}(e)). Subsequently, we perform a connected component analysis in 3D (Figure~\ref{fig:trackpipeline}(f)). By using these connected components together with the estimated camera poses from SfM, we can reproject the outlines of these connected components onto an image series. These outlines provided us with a series of Region of Interest (RoI) that all show a single cluster of fruits (Figure~\ref{fig:trackpipeline}(g)). A 3D cluster may appear in several frames (see Figure~\ref{fig:trackpipeline}(g)). We chose the three frames with the highest amount of segmented apple pixels and report the median count of these three frames as the fruit count for the cluster. 

\begin{figure*}[!hbpt]
	\centering
	\def\svgwidth{0.6\textwidth}
	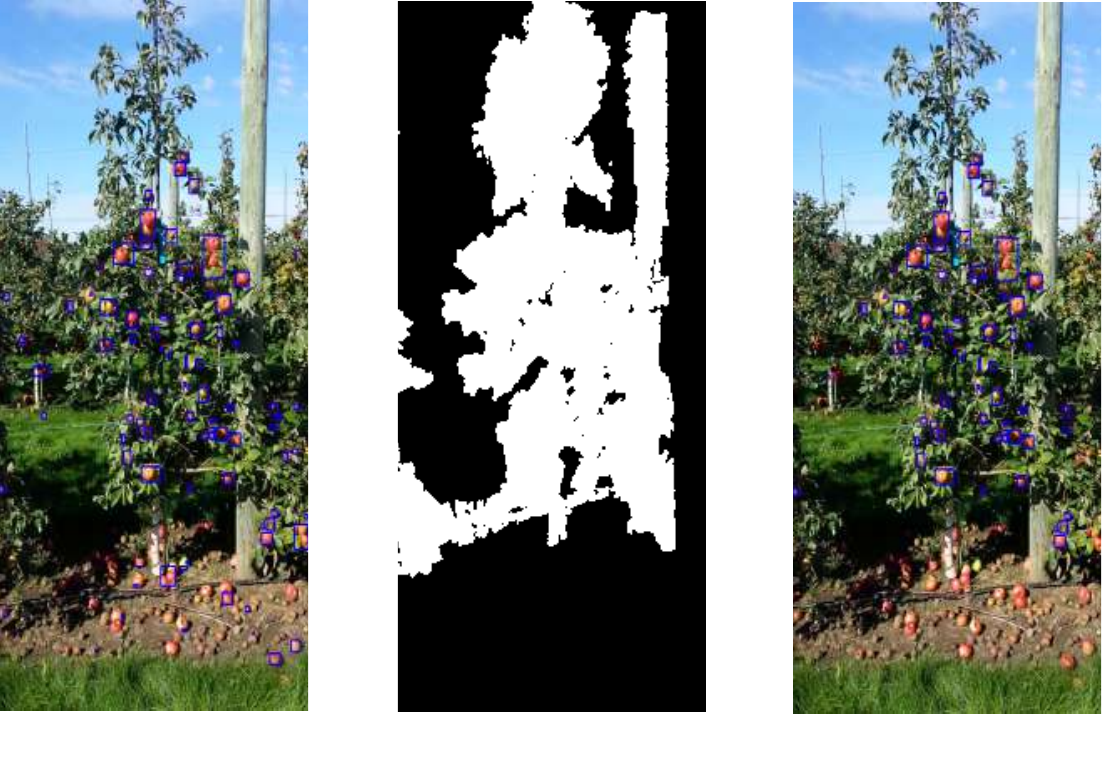\label{fig:gpremoved}
	\caption{Schematic overview of the removal of fruits on the ground and on trees in the background. (a) All Detections; (b) Mask for removing ground plane and fruits in the background; (c) Filtered detections}
	\label{fig:gp_removed}
\end{figure*}

We remove the apples on the ground and trees in the background for all the single-side counts by using a computed 3D ground plane and depth masks. We perform these steps for both sides of the row. Figure~\ref{fig:gp_removed} shows an overview of this process. Additionally, we need to merge the 3D reconstructions of the front- and back side of the tree row to eliminate double counting of fruits that are visible from both sides. We use the algorithm described in~\cite{roy_registering_2018, dong_semantic_2018} to accomplish this task. To merge fruit counts from both sides, we compute the intersection of the connected components from both sides (Figure~\ref{fig:trackpipeline}(h)). Then, we compute the total counts by summing up the counts from all the connected components, computing the intersection area among them (among 1, 2,..., the total number of intersecting clusters) and adding/subtracting the weighted parts using the Inclusion-Exclusion principle~\cite{andreescu_inclusion-exclusion_2004} (Figure~\ref{fig:trackpipeline}(i)). This method is thoroughly evaluated in Section~\ref{subsec:yieldMappingexp}.

\section{Datasets}
\begin{figure}[!htb]
	\centering
	\def\svgwidth{225bp}
	\subfloat[Tree rows used for training, data captured in 2015\label{fig:train}]{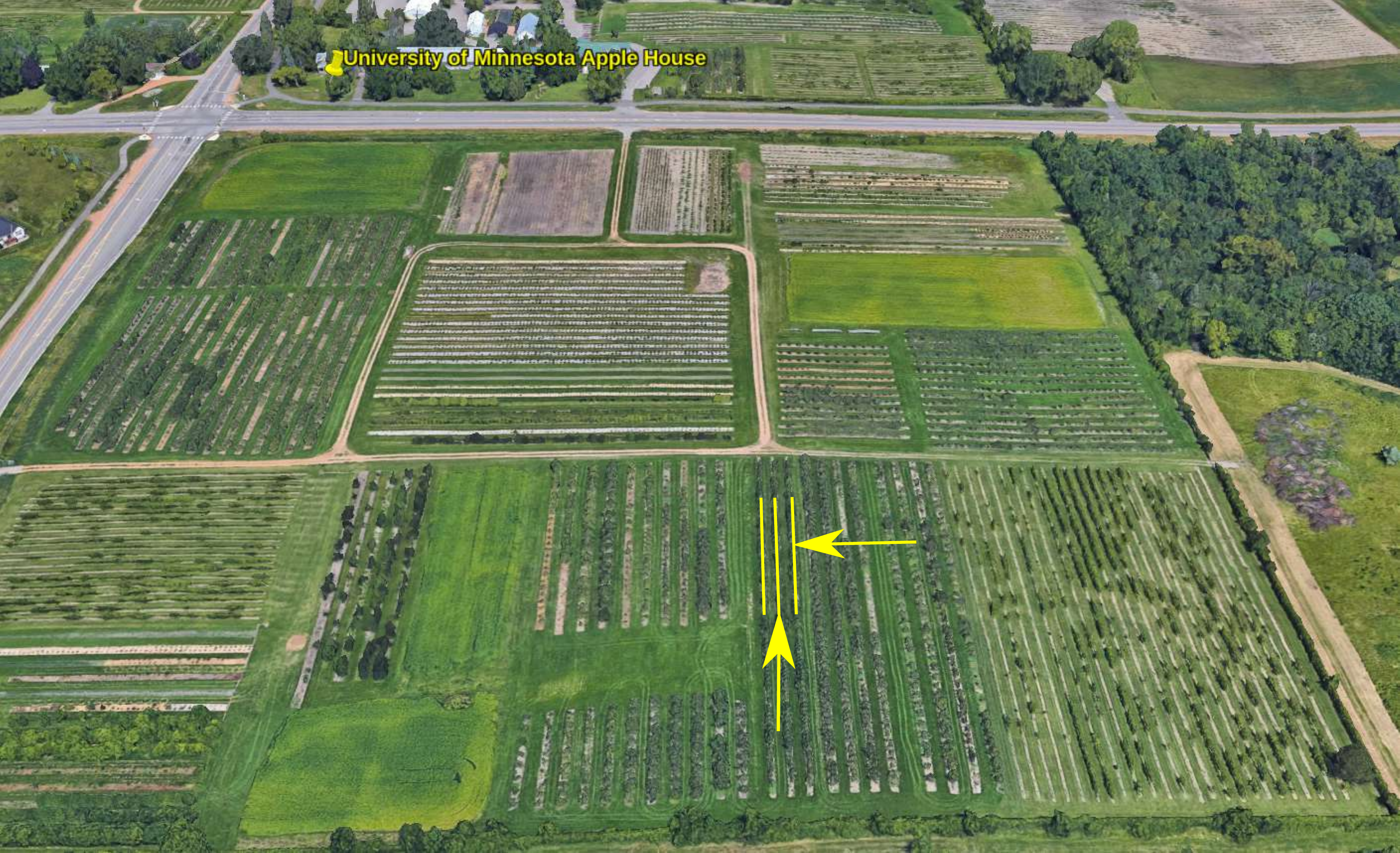}
	\quad
	\def\svgwidth{225bp}
	\subfloat[Tree rows used for testing, data captured in 2016\label{fig:test}]{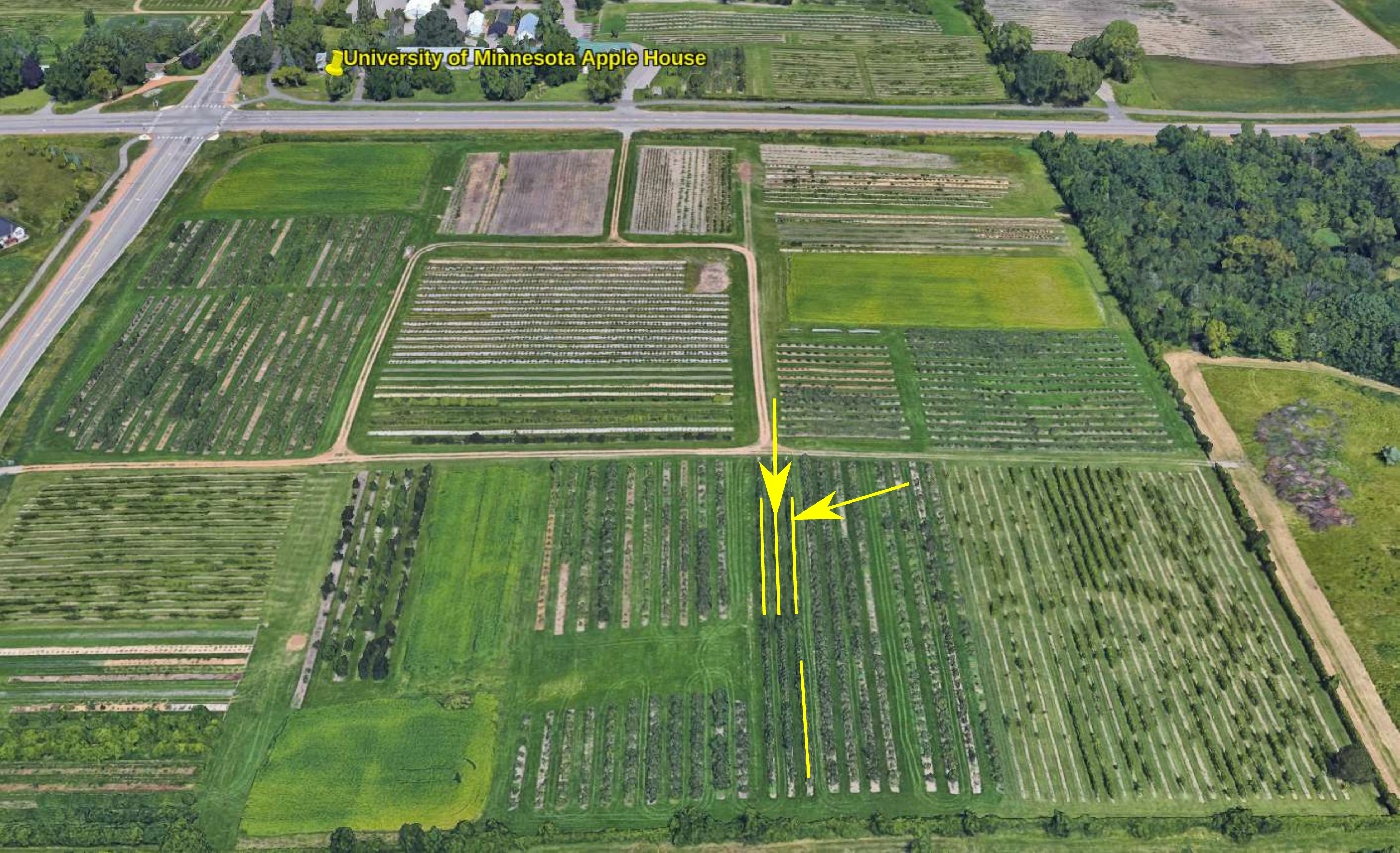}
	\caption{Tree row layouts at the University of Minnesota HRC}
	\label{fig:overview}
\end{figure}

All of the data used for this paper were collected at the University of Minnesota Horticultural Research Center (HRC) in Eden Prairie, Minnesota between June 2015 and September 2016. Since this is a university orchard, used for phenotyping research, it is home to a large variety of apple tree species. We collected video footage from different sections of the orchard using a standard Samsung Galaxy S4 cell phone. During data collection, video footage was acquired by facing the camera horizontally at a single side of a tree row. Individual images were extracted from these video sequences. Figure~\ref{fig:overview} shows the orchard layout and the tagged tree rows. Datasets $1-3$ use the same tree rows (same apple variety) for training and testing. However, the data was acquired in different years and during different growth stages.

\subsection{Datasets for Apple Detection / Yield Estimation}
\begin{figure}[!htb]
	\centering
	\subfloat{{\includegraphics[width=0.18\textwidth]{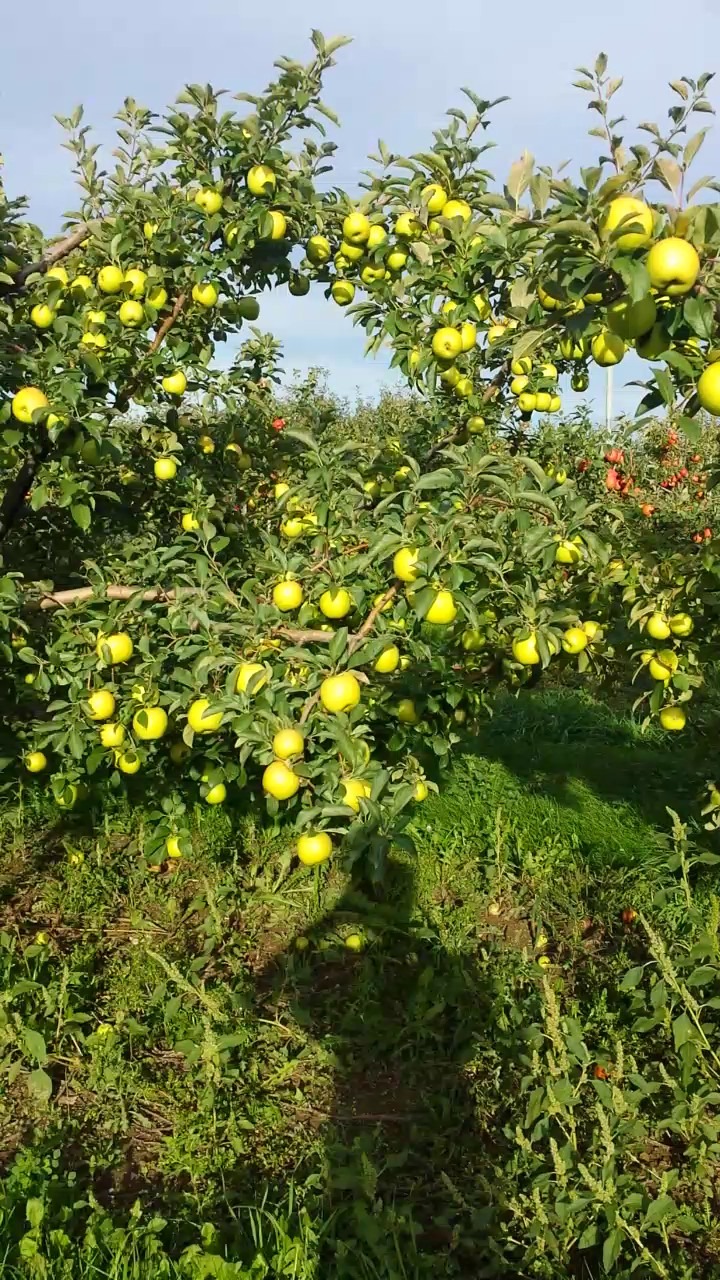}}}%
	\hspace{.2cm}
	\subfloat{{\includegraphics[width=0.18\textwidth]{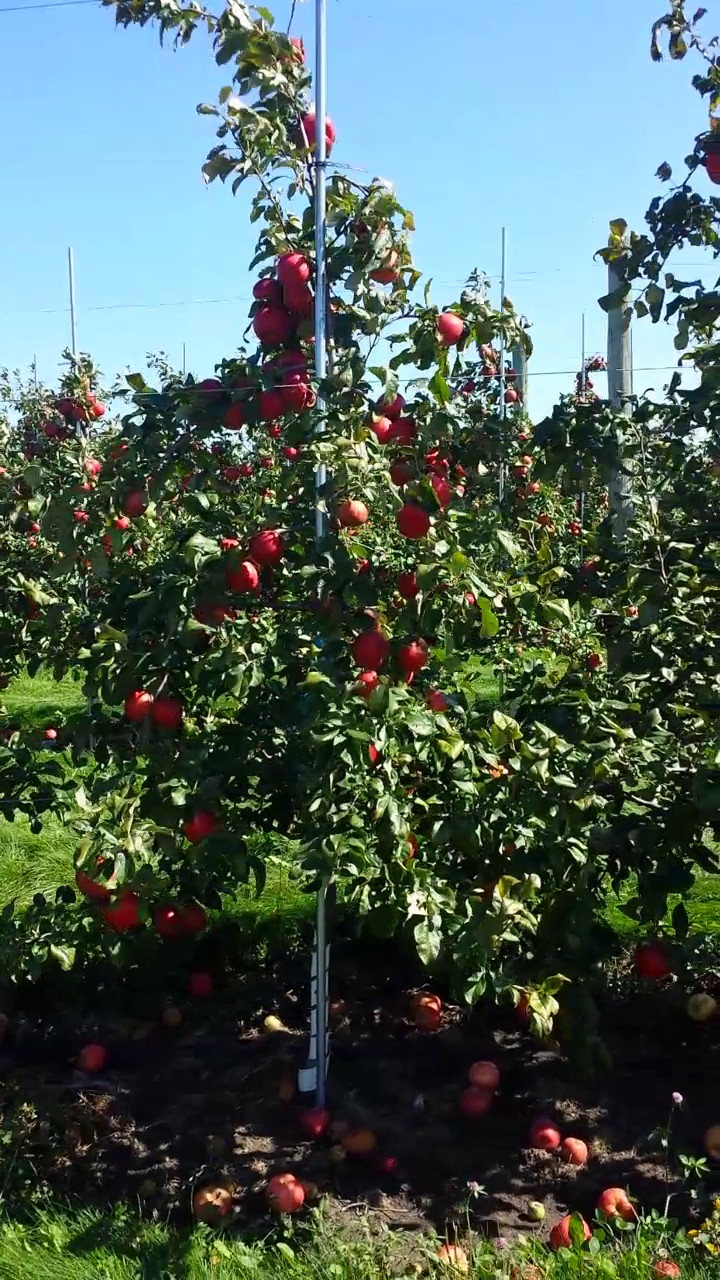}}}%
	\hspace{.2cm}
	\subfloat{{\includegraphics[width=0.18\textwidth]{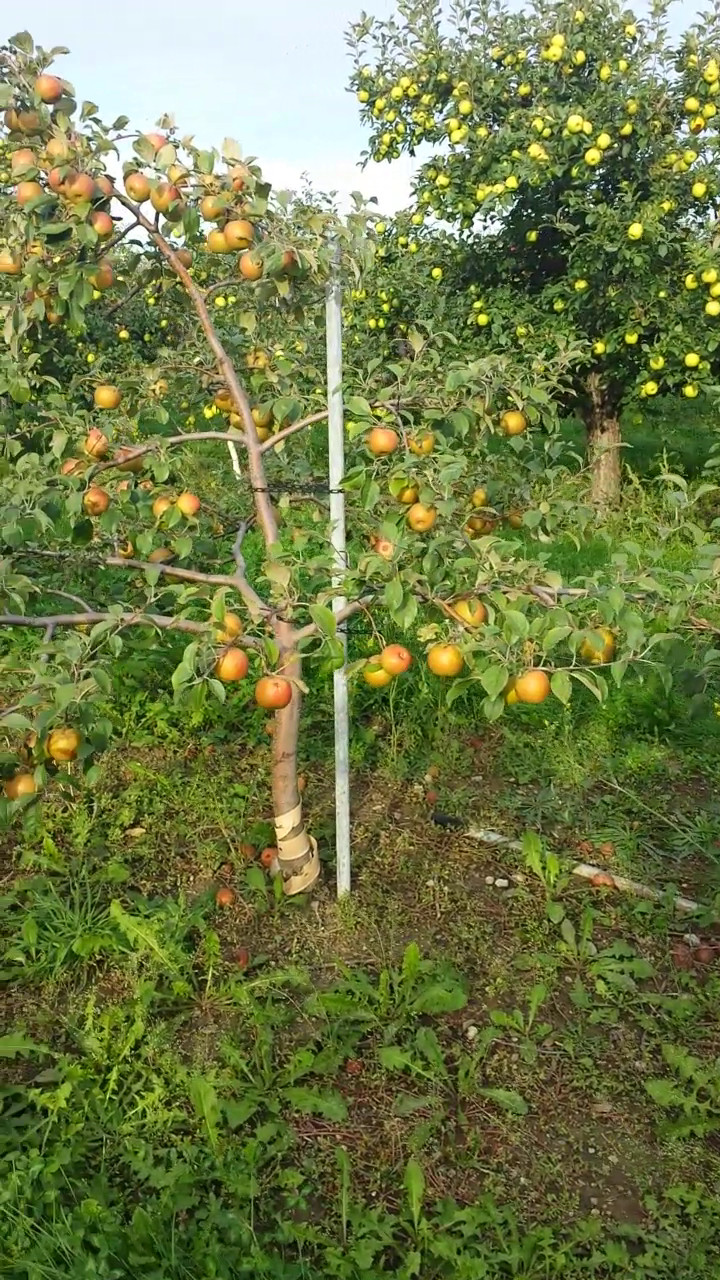}}}%
	\hspace{.2cm}
	\subfloat{{\includegraphics[width=0.18\textwidth]{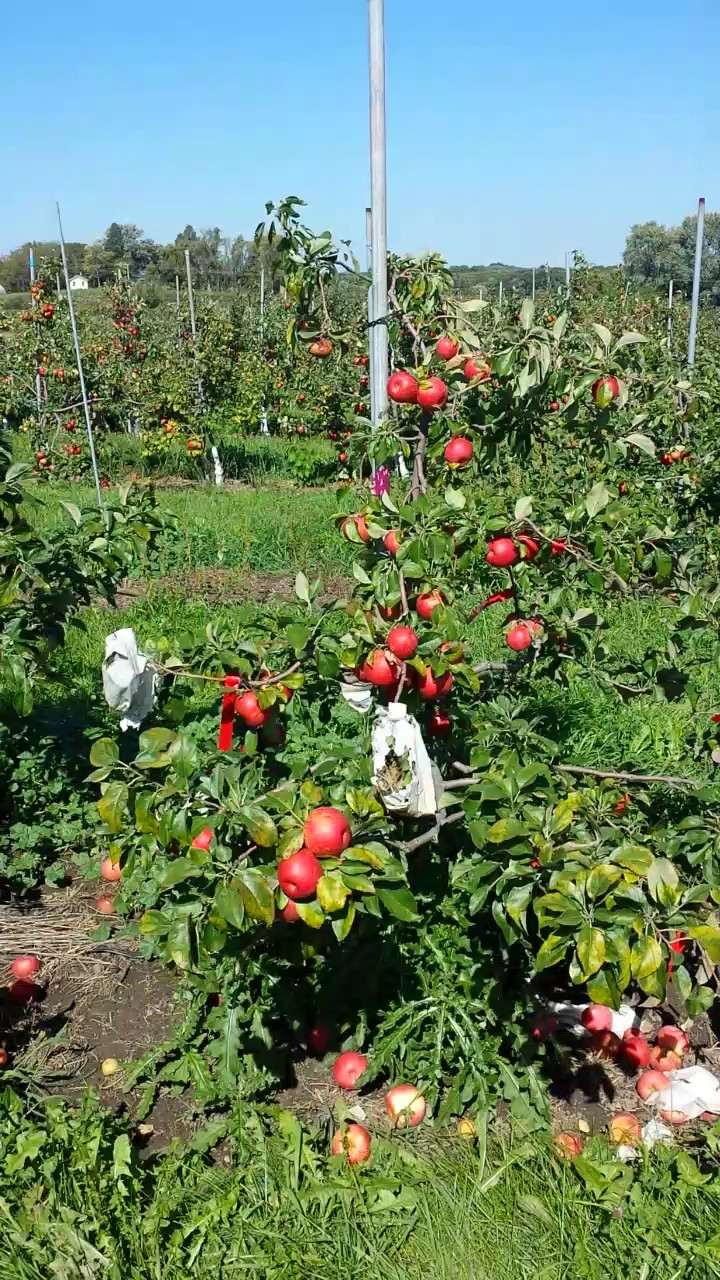}}}%
	\hspace{.2cm}
	\subfloat{{\includegraphics[width=0.18\textwidth]{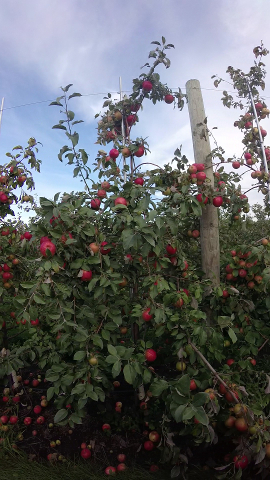}}}%
	\caption{Four sample images from the annotated datasets used for training U-Net and FRCNN (first four from left) and a sample image from the Semi-supervised GMM dataset (rightmost). }
	\label{fig:trainingsets}
\end{figure}

\textbf{Training Sets:} A total of $10$ datasets were sampled from $6$ tree rows (see Figure~\ref{fig:train}) for training the U-Net/FRCNN detection models. From these $10$ datasets we annotated $103$ images of size $1920 \times 1080$ pixels. All of these datasets were acquired in $2015$ at the HRC, and they contain different apple varieties, fruits across different growing stages and a variety of tree shapes. See Figure~\ref{fig:trainingsets}(first four from left) for a few sample images from the training sets.
To develop a training model for the Gaussian Mixture Model (GMM) based detection without user supervision, we collected an additional dataset in 2015, from the same row of the training set 2. We captured a video from a single side of the row using a Garmin VR camera. Instead of annotating the video manually, we obtained user supervisions in the form of clicks on apples for fifty frames, randomly sampled from the entire video. We refer to this dataset as the "Semi-supervised GMM" dataset for the rest of the paper. See Figure~\ref{fig:trainingsets}(rightmost) for a sample image from this dataset.

\textbf{Validation Sets:}
The validation data were sampled from the same $10$ datasets as the training data in the following fashion:
\begin{itemize}
	\item[] \textbf{GMM:} The GMM based method does not require annotated validation data.
	
	\textbf{U-Net:} We chose an $80/20$ split of the extracted patches for training and validation.
	
	\textbf{FRCNN:} We chose an $80/20$ split of the extracted patches for training and validation.
\end{itemize}

\begin{figure}[ht]
	\centering
	\subfloat[Dataset1]{\label{fig:valida}{\includegraphics[width=0.23\textwidth]{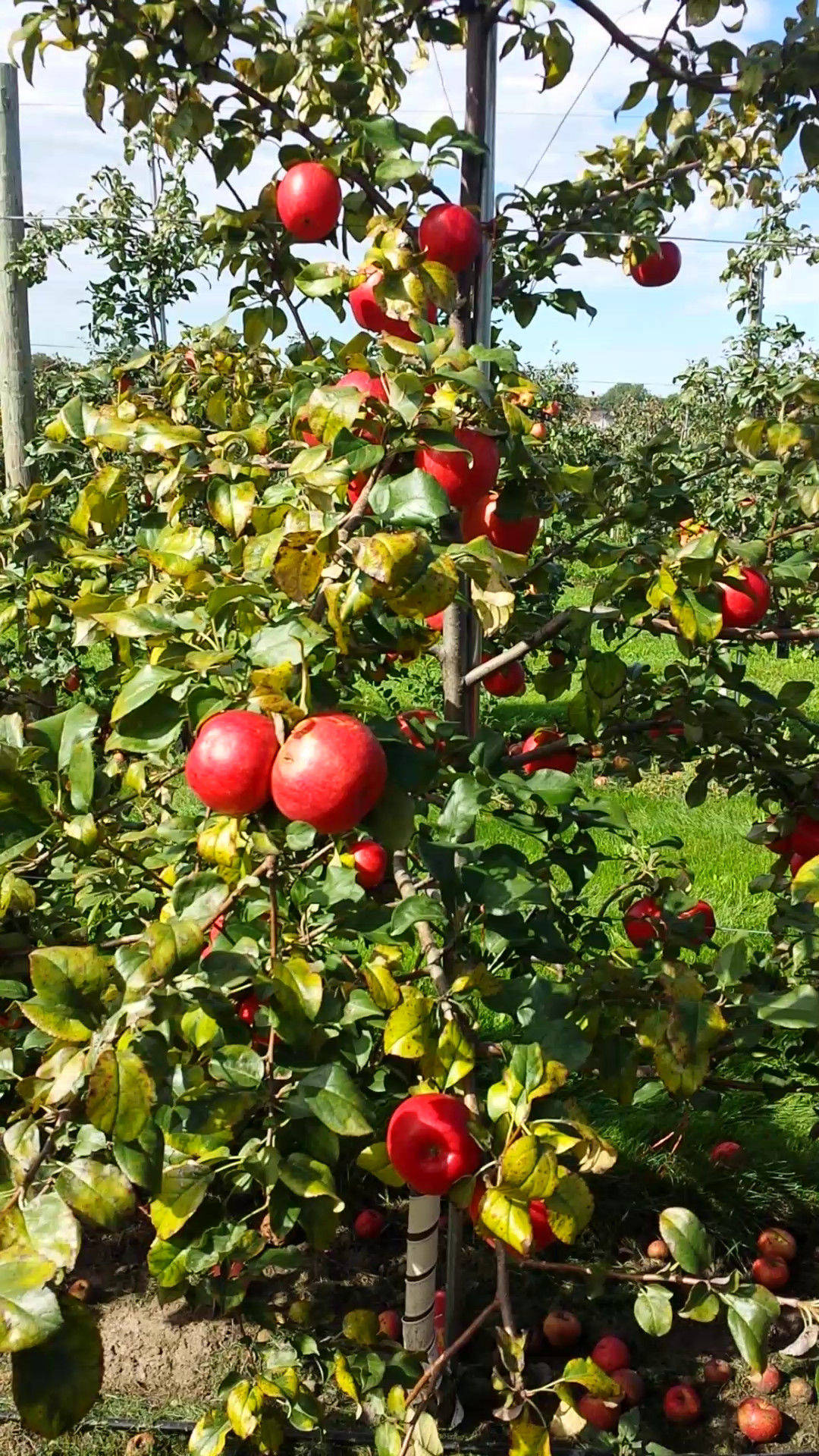}}}%
	\quad
	\subfloat[Dataset 2]{\label{fig:validb}{\includegraphics[width=0.23\textwidth]{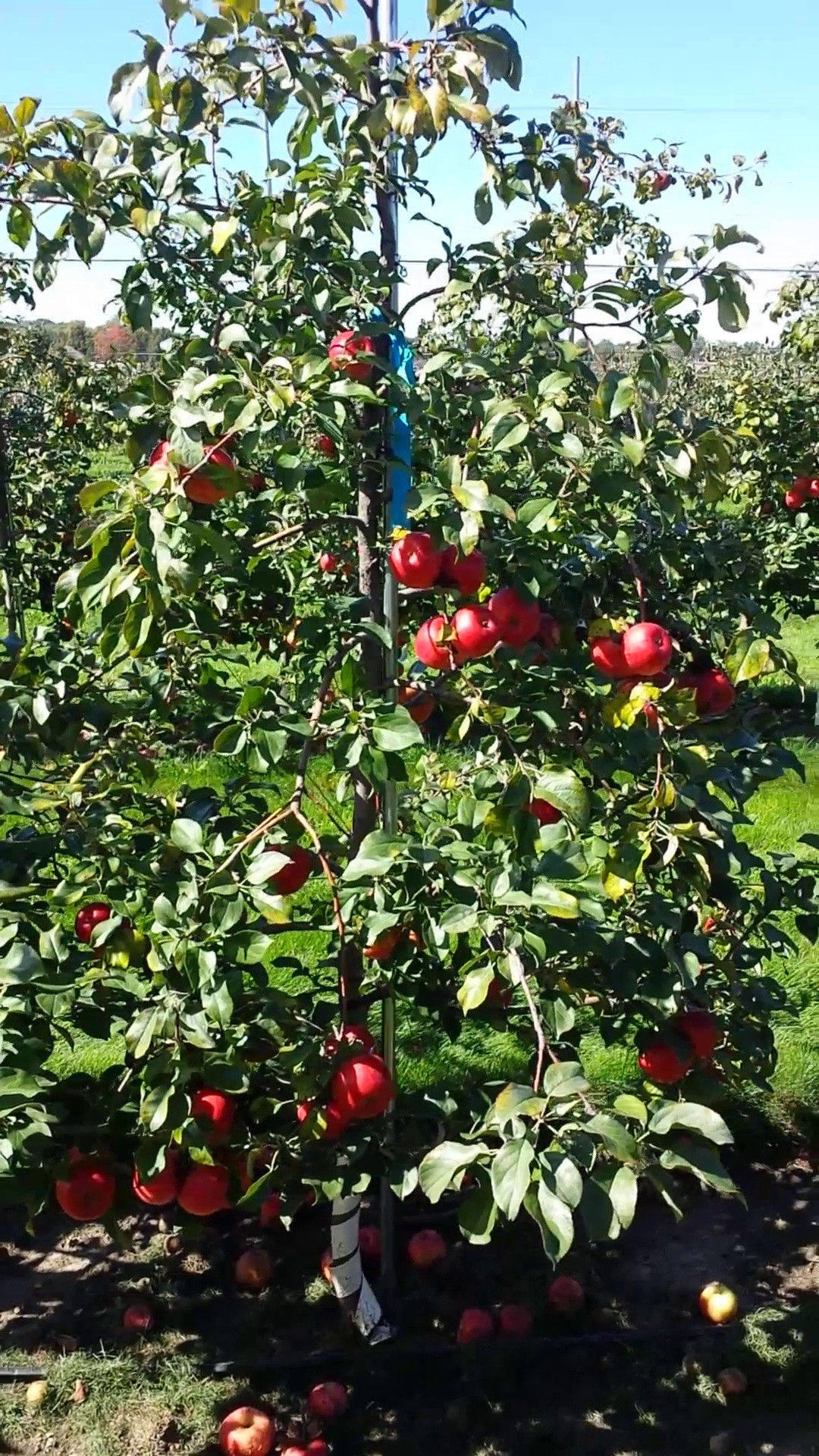}}}%
	\quad
	\subfloat[Dataset 3]{\label{fig:validc}{\includegraphics[width=0.23\textwidth]{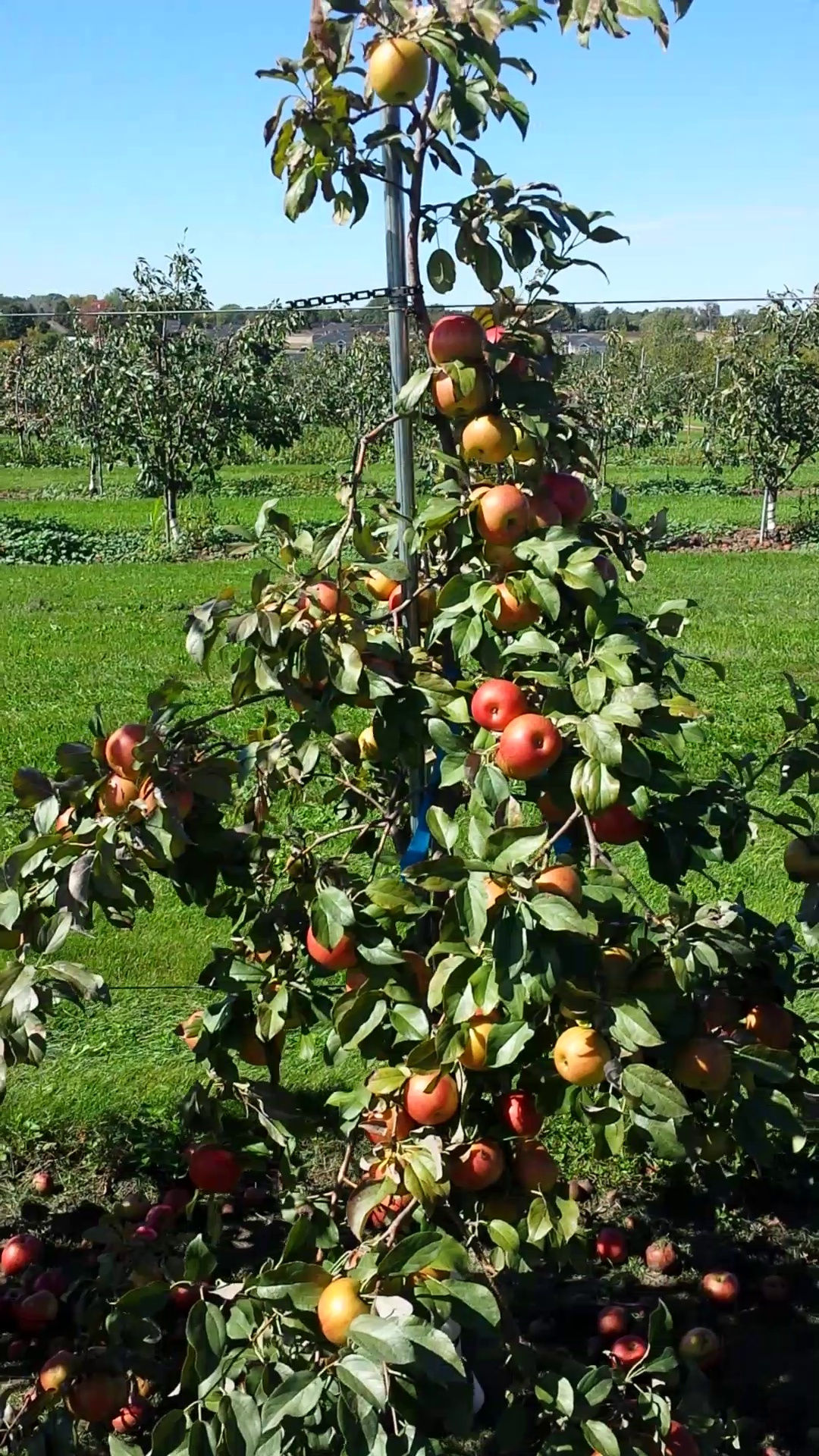}}}%
	\quad
	\subfloat[Dataset 4]{\label{fig:validd}{\includegraphics[width=0.23\textwidth]{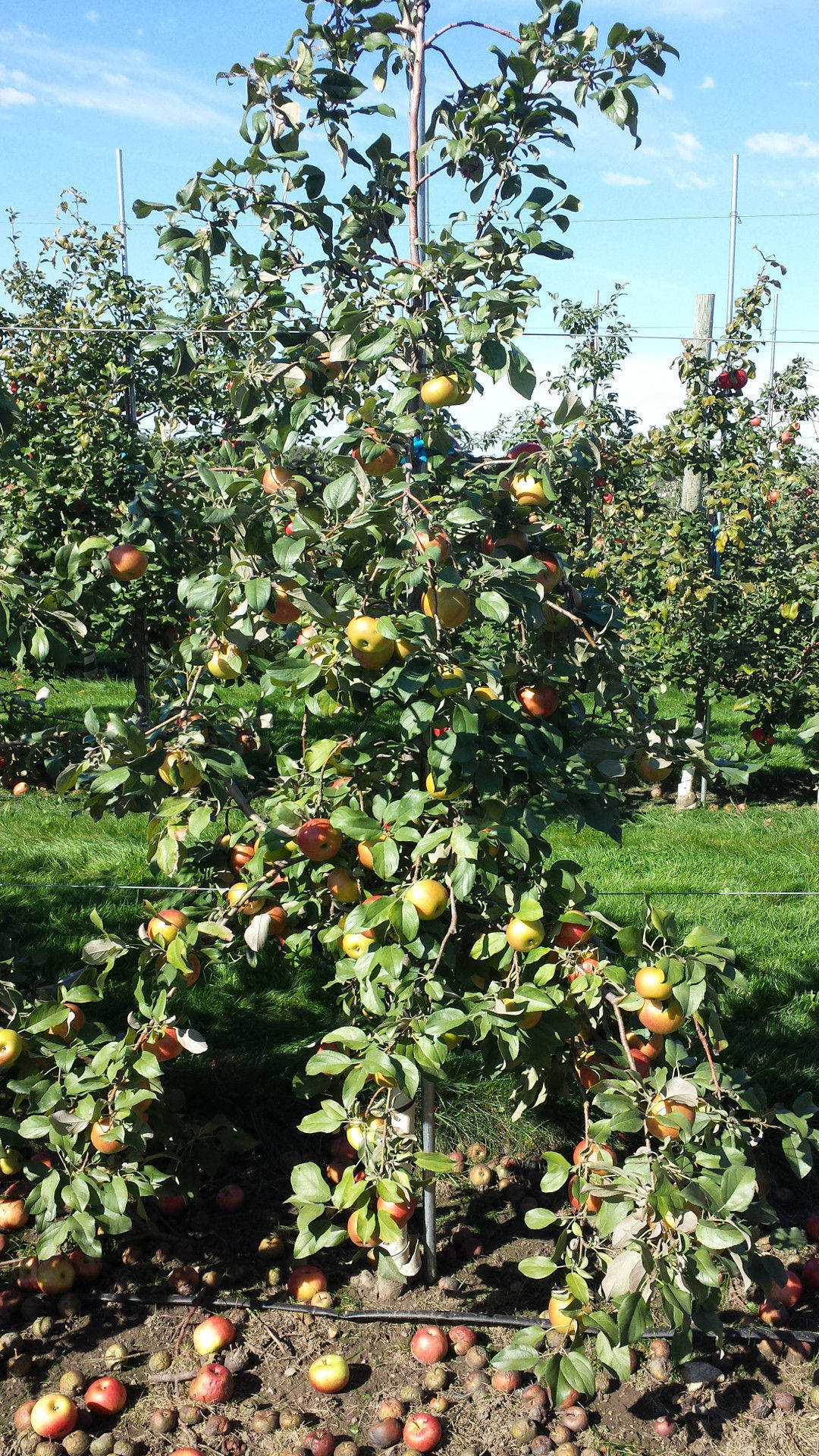}}}%
	\caption{Example images of the datasets used for testing of the detection and yield estimation stages}
	\label{fig:testsets}
\end{figure}

\textbf{Test Sets:}
To evaluate detection and yield estimation performance we arbitrarily chose four different sections of the orchard (see Figure~\ref{fig:test}). We collected seven videos from these four segments in 2016. These sections were annotated manually with bounding boxes to mark fruit locations. We also collected ground truth for the rows in question by collecting per tree yield and by measuring fruit diameters after harvest. We used all seven videos for the evaluation of the detection and videos from both sides of datasets 1, 2 and 3 for the yield estimation experiments. See Table~\ref{tab:data} for dataset details and see example images in Figure~\ref{fig:testsets}.

\begin{table}[ht!]
	\begin{center}
		\caption{Overview of the apple detection/yield estimation test datasets}
		\label{tab:data}
		\begin{tabular}{|c|c|c|c|}
			\hline
			\textbf{Dataset} & \specialcell{\textbf{Number} \\ \textbf{of trees}} & \specialcell{\textbf{Number of} \\ \textbf{harvested apples}}  & \textbf{Characteristics} \\
			\hline
			1 & 6 & 270 & \specialcell{Red apples, \\ planar geometry (apples are visible from both sides), \\ acquired late in the season (yellow leaves)}\\
			\hline
			2 & 10 & 274 & \specialcell{Red apples, \\ non planar geometry} \\
			\hline
			3 & 6 & 414 & \specialcell{Mixture of red and green apples, \\ non planar geometry} \\
			\hline
			4 & 4 & 568 & \specialcell{Mixture of red and green apples, \\ non planar geometry \\ Only collected video from the sunny side} \\
			\hline
		\end{tabular}
	\end{center}
\end{table}

\subsection{Datasets for Apple Counting}
\textbf{Training Sets: }For training of the cluster counting network, we used the same two datasets as in~\cite{hani_apple_2018}. One of these datasets contains green, and one contains red apples. They were acquired from same tree rows as datasets 5 and 6 (see Figure~\ref{fig:train}) in 2015. Both datasets were obtained from the sunny side of the tree row. From these two datasets, we extracted image patches using the GMM detection method described in Section~\ref{subsec:gmmdetection}. In total, we obtained $13000$ image patches, which were annotated manually. Additionally, we extracted $4500$ patches at random that do not contain apples. To balance our training dataset between classes we up-sampled the training dataset, using random data augmentation (horizontal flipping, rotations of $\pm 5^\circ$ and Gaussian smoothing), to a total of $\sim65,000$ image patches. 

\textbf{Validation Sets: }To supervise the network during training, we used n $80/20 \%$ split of the available data for training/validation.

\textbf{Test Sets: } To validate our patch based counting approach we used four datasets. Compared to the preliminary version presented in~\cite{hani_apple_2018}, we modified the data to (1) Subsample the first test dataset. This dataset previously contained 7 times more images than the other three datasets. (2) Remove patches that show apples lying on the ground. The final datasets are composed of a total of $2874$ images. See Table~\ref{tab:datacount} for dataset details.

\begin{table}[ht!]
	\begin{center}
		\caption{Overview of the apple counting test datasets}
		\label{tab:datacount}
		\begin{tabular}{|c|c|c|}
			\hline
			\textbf{Dataset} & \specialcell{\textbf{Number of} \\ \textbf{image patches}}  & \textbf{Characteristics} \\
			\hline
			1 & 956 & \specialcell{Red apples, \\ contains patches from the sunny and shady side of the tree row}\\
			\hline
			2 & 628 & \specialcell{Yellow and orange apples, \\ contains patches from the sunny side of the tree row} \\
			\hline
			3 & 587 & \specialcell{Green apples, \\ contains patches from the sunny and shady side of the tree row} \\
			\hline
			4 & 703 & \specialcell{Red apples, \\ contains patches from the sunny side of the tree row \\ acquired from larger distance (apples appear at lower resolution)} \\
			\hline
		\end{tabular}
	\end{center}
\end{table}

\subsection{Manual Annotation of Datasets}
\label{sec:labeling}
To validate our detection and counting methods, we need image level ground truth. The used annotation process differs slightly between training and validation datasets. 

\textbf{Detection Training Sets for U-Net and FRCNN:} The fruit detection and yield estimation training sets were annotated using the VGG annotator tool~\cite{dutta_vgg_2016}. Apples on the foreground trees were annotated using polygons. Apples on the ground and trees in the background were not tagged.

\textbf{Detection Training Sets for GMM:} For the semi-supervised and user-supervised GMM the images were over-segmented into $25$ color clusters. Then users were provided with an interface where they can click on the fruits. The closest color cluster to the selected pixel color was found, and the rest of the pixels belonging to this color class were shown to the users. Users added color clusters to the model in this fashion until most of the fruits were identified.

\textbf{Detection Test Sets:} Fruits in these datasets were annotated using bounding box annotations. For manual annotation, frames were selected arbitrarily every $1$ to $3$ second from the test videos (frame rate 30 fps), depending on how much the camera moved since the last annotated frame. Apples on the ground and trees in the background were not tagged.

\textbf{Counting Training and Test Sets: } Both counting methods for this paper were evaluated on small image patches. The extracted image patches were annotated by hand with a single ground truth $\text{count} \in [0,6]$. At least two human labelers annotated each image. Discrepancies were resolved by a third inspection of the image patches in question.
\section{Experiments and Results}
\label{sec:experiments}
In this section, we evaluate each of the presented methods for fruit detection and counting quantitatively. Additionally, we give some qualitative insights and analyze some common failure cases.

\subsection{Training Procedure}
\textbf{Training the GMM:} To evaluate the GMM based detection method with and without user supervision, we developed two models. To evaluate the performance without user supervision, we use the semi-supervised GMM dataset. This dataset was captured on a different year with a different camera. We train the model in a supervised fashion, by adding relevant clusters. The user-supervised model is trained by clicking on apples for the first five frames for each test video.

\textbf{Training U-Net:} For training the U-Net, we extracted patches of size $224 \times 224$ pixels with a stride of $50$ pixels from the original images. In total, $\sim 59000$ annotated patches were extracted for training.

\textbf{Training FRCNN:} Our implementation of Faster R-CNN loosely follows~\cite{bargoti_deep_2017}. Initially, our network was trained on the same data, which is available open source~\footnote{\url{http://data.acfr.usyd.edu.au/ag/treecrops/2016-multifruit/}}. This data contains a total of $1120$ images with $308 \times 202$ pixels resolution. We used the same train/validation split like the one used in their paper. To offer a fair comparison to our proposed method, we added our own annotated data to the training set. We extracted image patches of size $500 \times 500$ pixels by moving a sliding window with stride $50$ pixels over the annotated images. In total, $\sim 9000$ annotated patches were extracted for training.

\subsection{Detection Results}\label{subsec:detection_result}
\begin{figure}[htbp!]
	\centering
	\includegraphics[width=\textwidth]{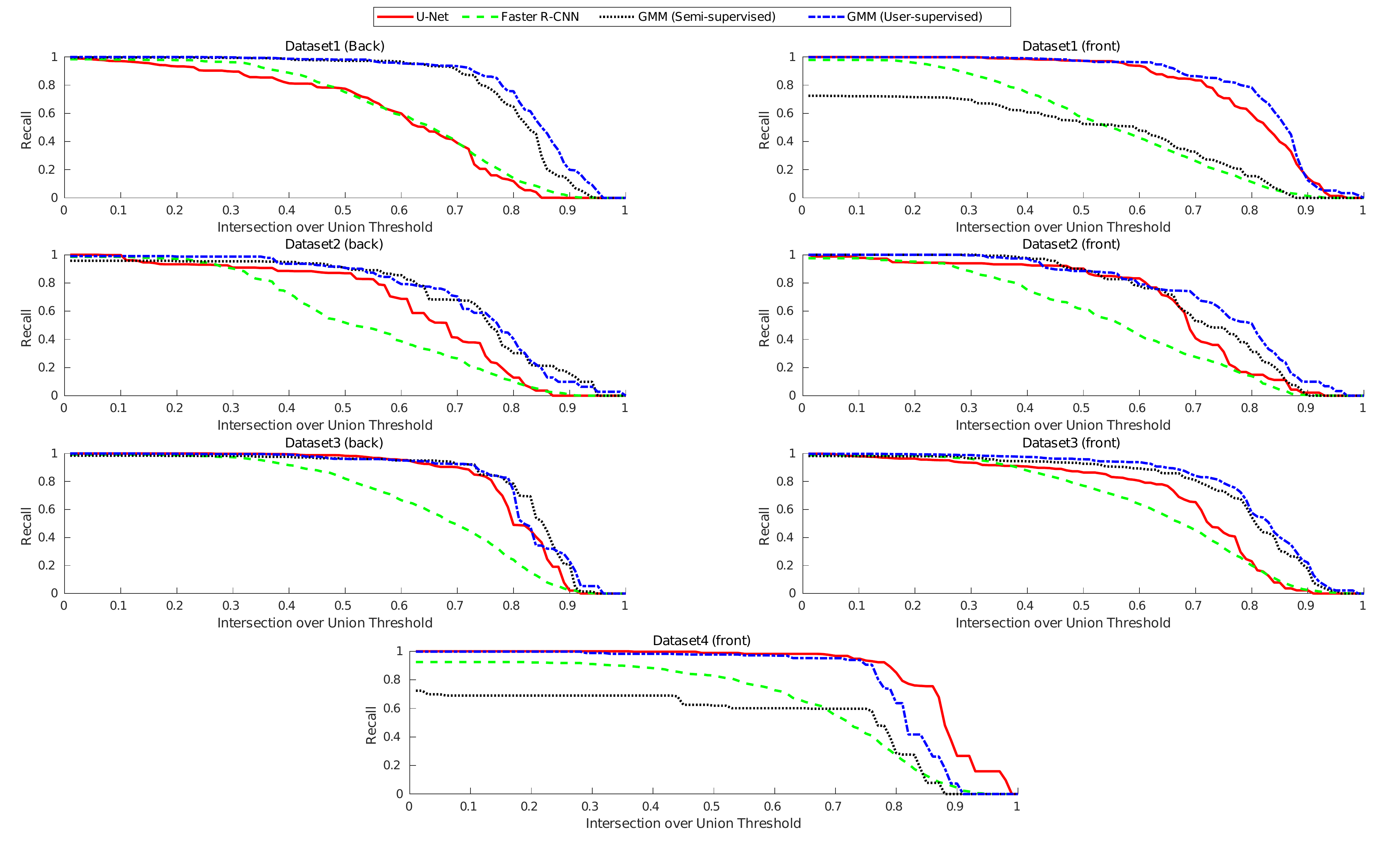}
	\caption{Recall of the detection methods on all seven datasets}
	\label{fig:recall}
\end{figure}

We use three metrics for evaluation purposes: precision, recall and $F_1$-measure. These metrics are obtained using true positives (TP), false positives (FP) and false negatives (FN) rates. Recall is a measure of how many relevant objects are selected out of the total number of objects. Formally, $\text{Recall} = \dfrac{TP}{TP + FN}$. Precision is a measure of how many of the detected objects are relevant. Formally, $\text{Precision} = \dfrac{TP}{TP + FP}$.
The $F_1$-measure is the harmonic mean between precision and recall. Formally, $\text{F1} = 2 \left(\dfrac{\text{Precision} \cdot \text{Recall}}{\text{Precision} + \text{Recall}}\right)$.
We computed these three measures for all seven test datasets over the entire Intersection over Union (IoU) range ($[0.01,0.99]$). IoU is defined as the area of overlap between detection and an object instance, divided by the area of the union of the two. The metrics are computed per frame, and we report the average per dataset.
Figure~\ref{fig:recall} shows the recall of all the four detection methods on the seven datasets. 

The user-supervised GMM method outperforms the other three approaches on $6$ out of seven datasets and is competitive on the last one. This outcome is expected, as the training models were tuned for each test set. The reason it falls behind in Dataset4 (front) is that color features alone were not enough to detect all the fruits. The performance of the semi-supervised GMM depends on how closely the test set color space resembles the training model. In the case of Dataset1 (front) and Dataset4 (front) the color space of the test set was different from the training model and therefore the recall drops substantially. In the other five test sets, the fruit colors were similar to the training set and the model achieves similar recalls to the user-supervised model. 
Our proposed technique based on U-Net achieves consistently high recall for all the datasets. In Dataset4 (front) it even outperforms the user-supervised GMM. This success can be attributed to the use of non-color features. The FRCNN method also achieves high recall (especially in the low IoU region. This is consistent with \cite{bargoti_deep_2017}, who suggested to use the FRCNN with $IoU = 0.2$. 

\begin{figure}[!htbp]
	\centering
	\includegraphics[width=\textwidth]{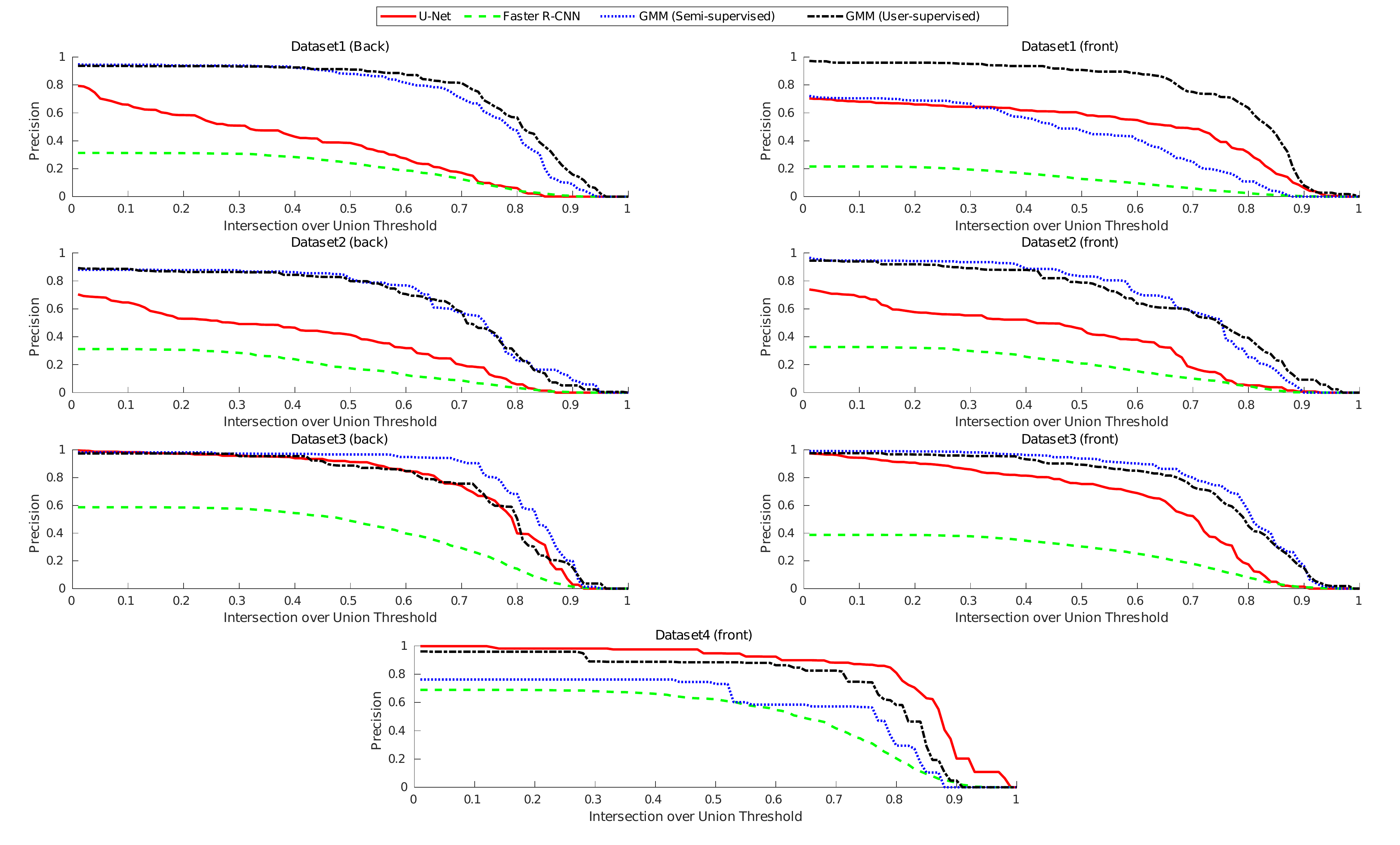}
	\caption{Precision of the detection methods on all seven datasets}
	\label{fig:precision}
\end{figure}

When we look at the plots showing the precision in Figure~\ref{fig:precision} the story is a different one. The user-supervised GMM approach outperforms the other three methods by a large margin in six out of seven test sets. The user-supervised GMM method is the only method to achieve precision values of $>90\%$ on all datasets. These high precision can be attributed to conservative user supervisions, which purposefully avoided ambiguous color clusters. The semi-supervised GMM achieves precision values of $>90\%$ in five out of seven datasets. For Dataset2 (back) and Dataset4 (front), the precision values fall under $80\%$ for all IoU levels. Again, this drop can be attributed to the dissimilarity in color space and lighting condition between the train and test sets. 

The U-Net based approach does not achieve precision values over $80\%$ in four out of seven datasets. We show some examples of kind of false positives in Section~\ref{sec:qualitative}. From the qualitative results, we see that the network detects yellowing leaves as apples. The apple trees did not have any yellowing leaves during 2015, and consequently, our training data did not have any such examples. However, this problem can likely be solved by adding more varied training data. When color features alone were not enough to detect all the fruits, as in Dataset4 (front), the network has higher precision than all other methods. 
The FRCNN method has the lowest precision on all datasets even with low IoU. This is due to three reasons: First, the network detects more false positives than the other two methods. Second, the FRCNN often merges separate object instances into one. Third, the Non-Maximum Suppression (NMS) either filters out true positives or returns multiple detections for a single object instance. NMS is prone to such behavior due to the manual choice of its threshold. In Section~\ref{sec:qualitative} we show qualitative examples that illustrate this behavior.  
These findings again confirm \cite{bargoti_deep_2017}. In their work, they estimated that roughly $4\%$ of the total error of their predictions could be attributed to the network's inability to distinguish clustered fruits. Instead, they treat detections as clusters, not only as single fruits, which helps increase precision. 

\begin{figure}[!htbp]
	\centering
	\includegraphics[width=\textwidth]{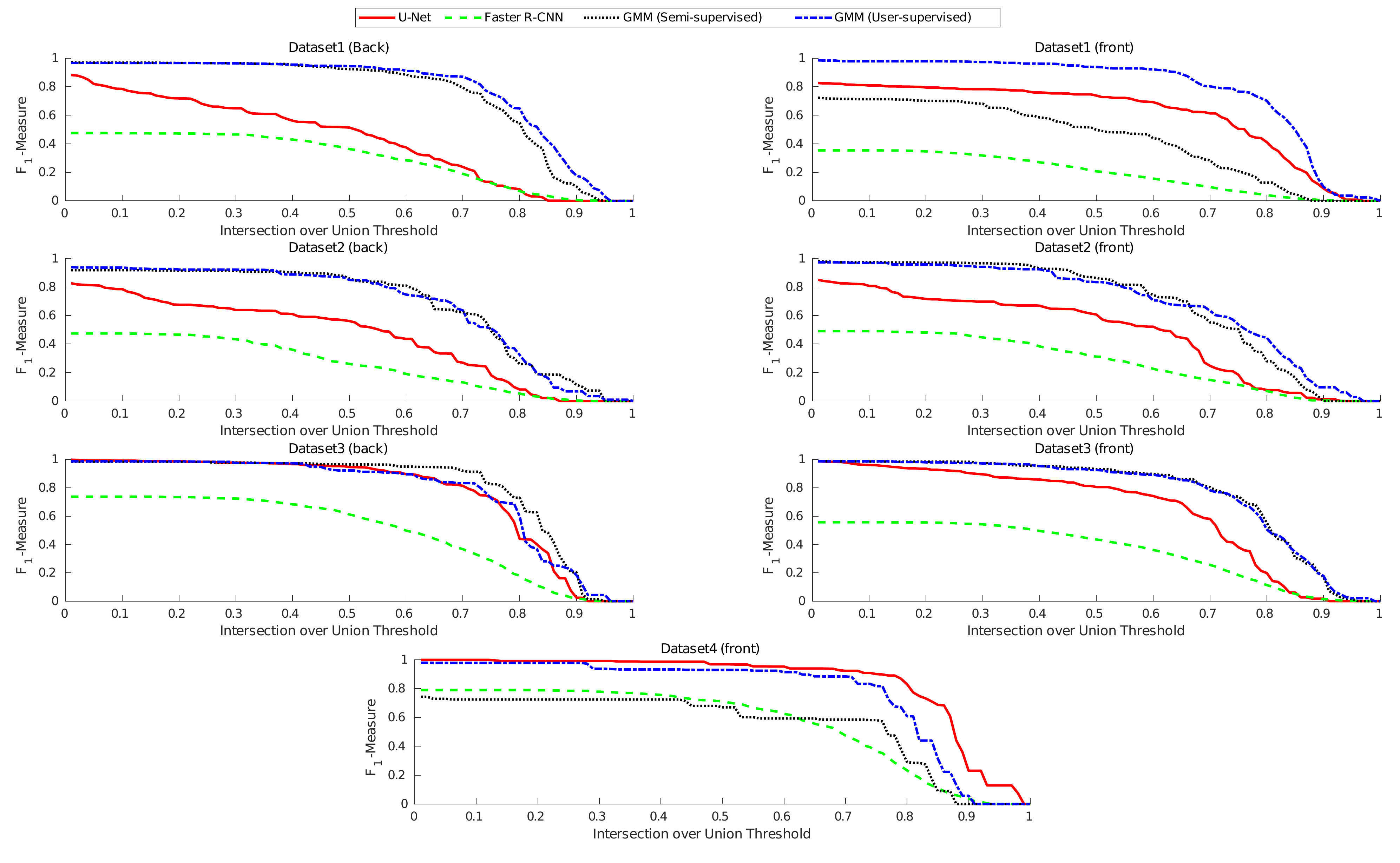}
	\caption{F1-measure of the detection methods on all seven datasets}
	\label{fig:F1}
\end{figure}

Since $F_1$-measure combines recall and precision, we expect the user-supervised GMM to outperform the other three approaches on most of the datasets as seen in Figure~\ref{fig:F1}. However, the U-Net detection approach achieves competitive $F_1$-measure on Dataset3 (back) and even outperforms the GMM approach on Dataset4 (front).

After these extensive evaluations, we can conclude with conviction that when fruits are distinguishable by colors, it is hard to beat the user-supervised GMM model. If color features are not good enough, we do need a more robust solution such as U-Net. However, the training set needs to cover a broader range of apple varieties, growth stages, and lighting conditions.

\subsubsection{Timing}
Detection performance is arguably the most important aspect of an object detection algorithm. A second aspect driving the choice of algorithm is its time complexity. The GMM approach runs at $5$ frames per second. The U-Net used in our experiments runs in less than 100 ms per image patch. We use image patches of size $224\times 224$ pixels with zero overlap. A full image of size $1920 \times 1080$ pixels takes less than $4.5$ seconds per frame.  For the FRCNN we follow the tiling approach proposed by~\cite{bargoti_deep_2017}. They use image crops of size $500 \times 500$ pixels with a stride of $50$ pixel. The FRCNN network takes $120$ ms per image patch. To detect objects on an image of $1920 \times 1080$ pixels takes up to $46$ seconds per frame. We acquire images at a rate of $30$ frames per second and move at a speed of $2$~m/s which renders the tiling approach infeasible for large orchards. These timings were measured on a conventional laptop with a single NVIDIA Quadro M1000 GPU.

\subsection{Counting Results}\label{subsec:count_result}
We repeated the experiments of counting apple clusters presented in~\cite{hani_apple_2018} with the modified network. The test datasets were adapted, so the results are not directly comparable. We evaluate counting performance of both methods discussed in Section~\ref{sec:counting}. Table~\ref{tab:1} shows the counting accuracy.

\begin{table}[h!]
	\begin{center}
		\caption{Image Patch Counting Results}
		\label{tab:1}
		\begin{tabular}{|c|c|c|c|c|}
			\hline
			\textbf{Approach} & \textbf{Test Set 1} & \textbf{Test Set 2} & \textbf{Test Set 3} & \textbf{Test Set 4}\\
			\hline
			GMM & 88.0 \% & 81.8 \% & 77.2 \% & 76.1 \% \\
			\hline
			ResNet50 & \textbf{88.8} \% & \textbf{92.68 \%} & \textbf{95.1 \%} & \textbf{88.5 \%}\\
			\hline
		\end{tabular}
	\end{center}
\end{table}

\begin{figure}[ht!]
	\centering
	\subfloat[GMM confusion matrix]{{\includegraphics[width=0.46\textwidth]{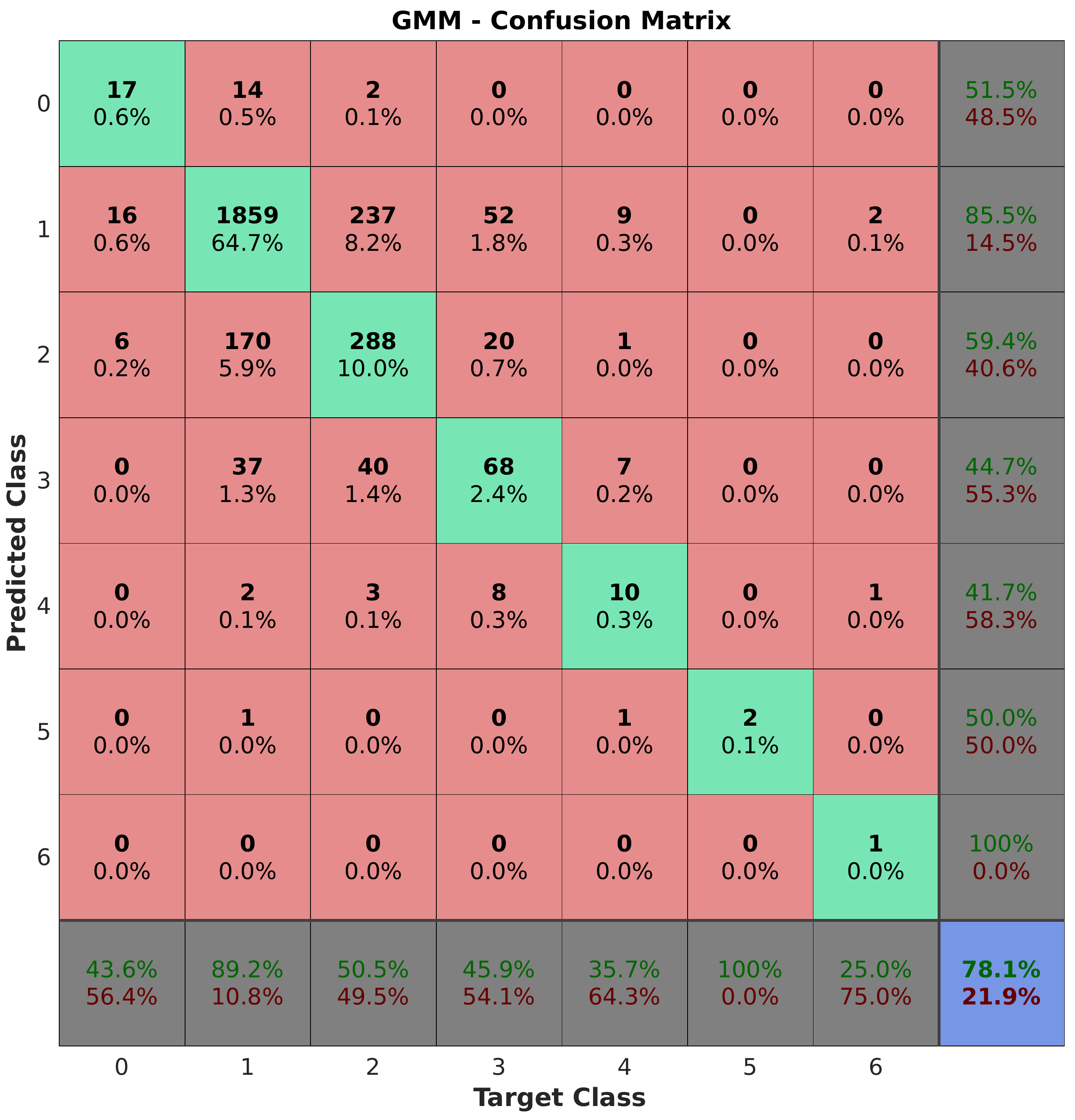}}}%
	\quad
	\subfloat[ResNet50 confusion matrix]{{\includegraphics[width=0.46\textwidth]{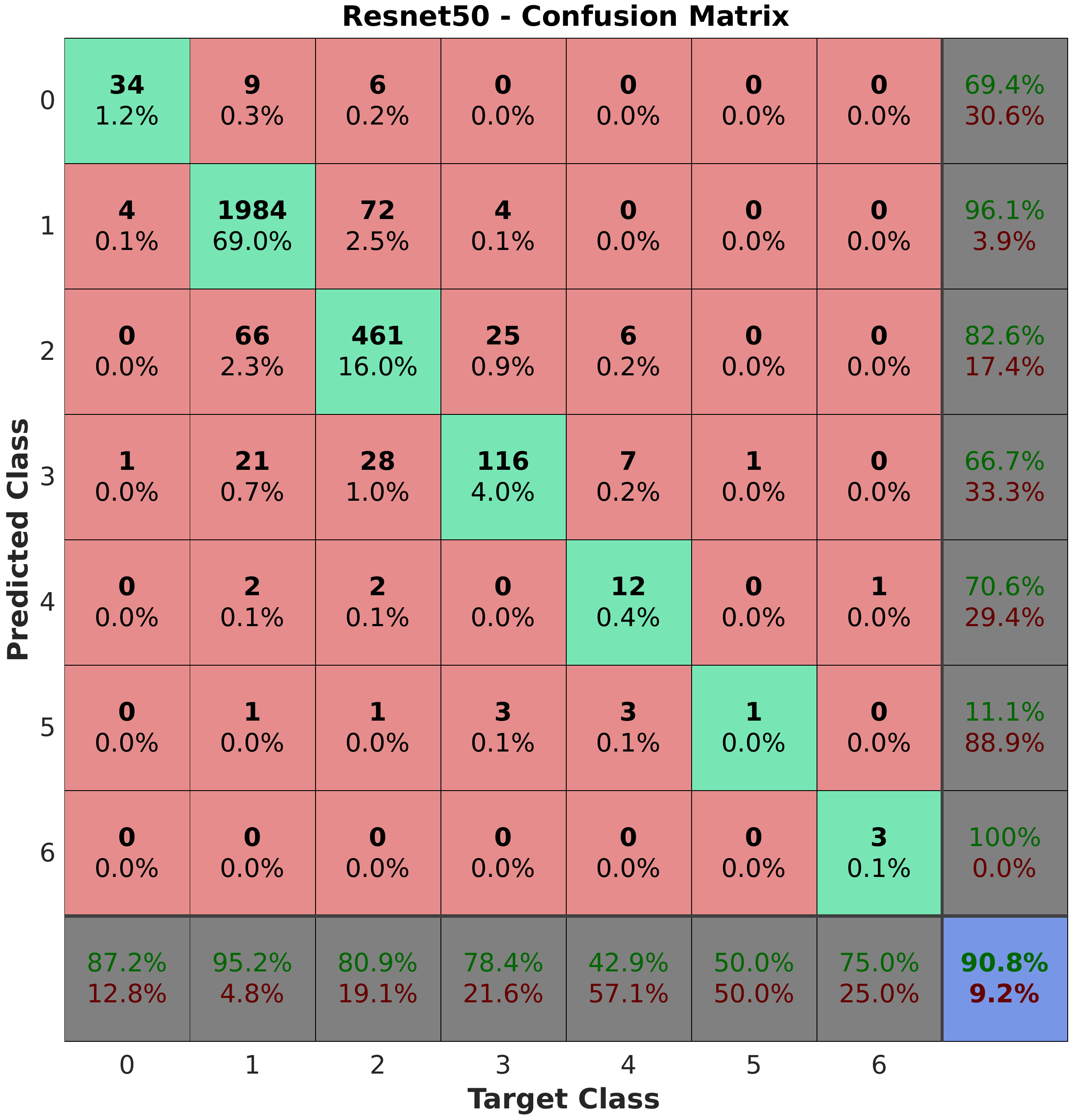}}}%
	\caption{Confusion matrices over all four test datasets}
	\label{fig:confmat}
\end{figure}

The ResNet50 network outperforms the GMM model on all of the test sets. On the test set 3 the CNN outperforms the GMM by almost $11\%$. On test sets 2 and 4 it outperforms the GMM by $18\%$ and $12\%$ respectively. These results show that the proposed neural network generalizes between datasets even under varying illumination conditions and among data sets with different colors.
The confusion matrices in Figure~\ref{fig:confmat} show that the GMM is precise in predicting a single apple. For all other categories, the performance of the method drops considerably. In contrast, the neural network's performance drops slowly in cases of higher fruit counts. Additionally, we see that the deep network can reject false positive detections in $87\%$ of the cases. Comparably, the GMM does so in only $43\%$ of the cases. The distribution of apples in these test datasets is highly skewed towards one single apple. We observed this phenomenon throughout our experiments, as the majority of clusters returned by the detection method are in the range $\lbrack 0, 4 \rbrack$. This finding suggests that our assumption of seven classes (fruit counts from $0$ to $6$) was too broad.

\subsection{Qualitative Results}
\label{sec:qualitative}
In this section, we show some qualitative examples from three datasets. We illustrate the performance of the three detection approaches on a sample image from each of these datasets. In Dataset4 (front) color features alone were not enough to detect all the apples; causing problems for the user-supervised GMM. Dataset1 (front) contains many yellowing leaves causing problems for both the U-Net and FRCNN. On Dataset3 (back) both the GMM and U-Net achieved high precision and recall, but the FRCNN still had poor precision.

\begin{figure*}[!hbpt]
	\centering
	\def\svgwidth{0.82\textwidth}
	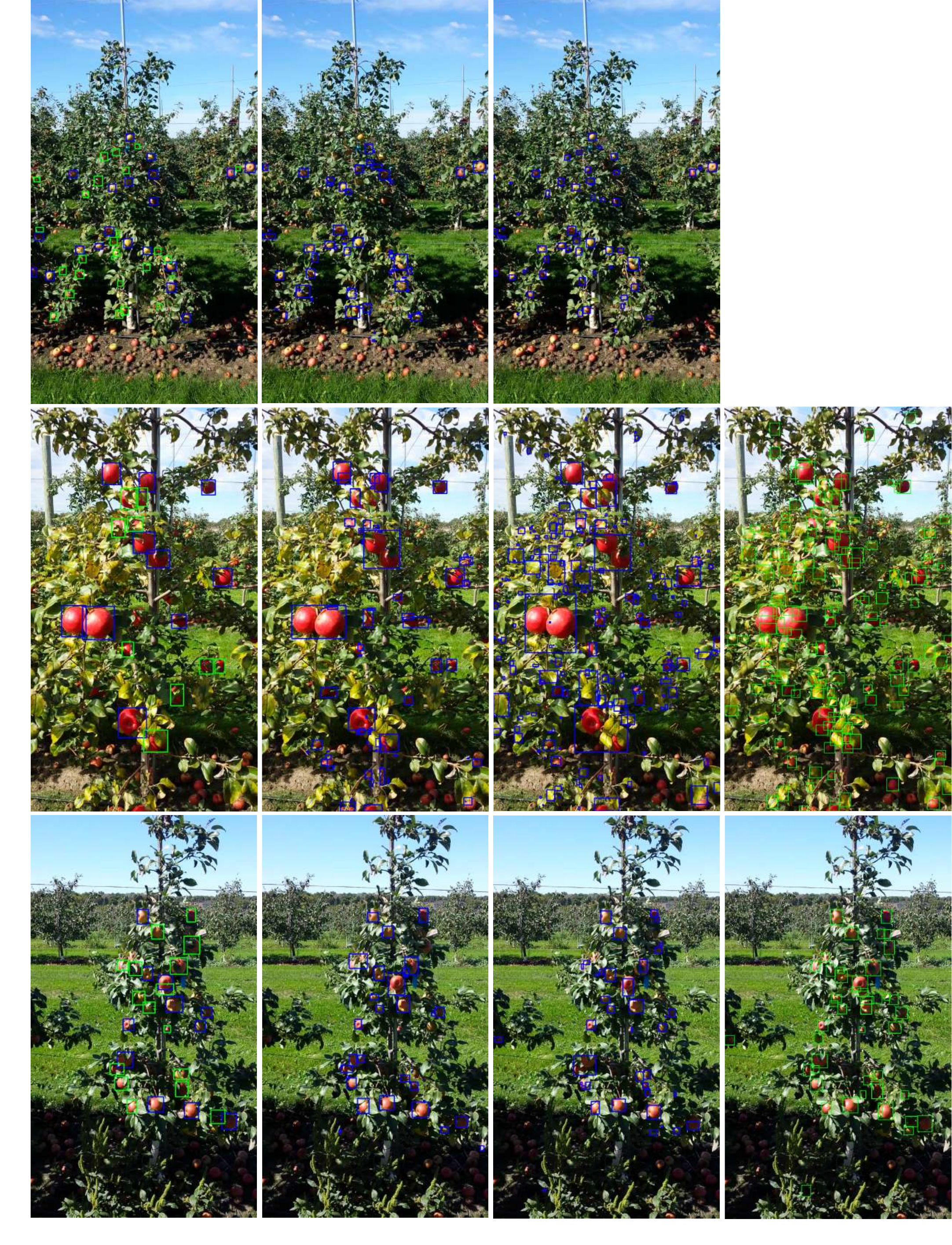
	\caption{Some qualitative results for the proposed detection methods}
	\label{fig:qualitative}
\end{figure*}

\subsection{Yield Estimation Results} \label{subsec:yieldMappingexp}
Detecting fruits and counting them in a per frame manner are both technically challenging tasks. However, these are only subproblems to the yield estimation task, which we ultimately want to solve. Our goal in this section is to find the best possible system using the previously discussed methods. 

\begin{figure}[htbp!]
	\centering
	\subfloat[][Yield mapping by summing up fruit counts from individual sides]{\includegraphics[width=0.5\textwidth]{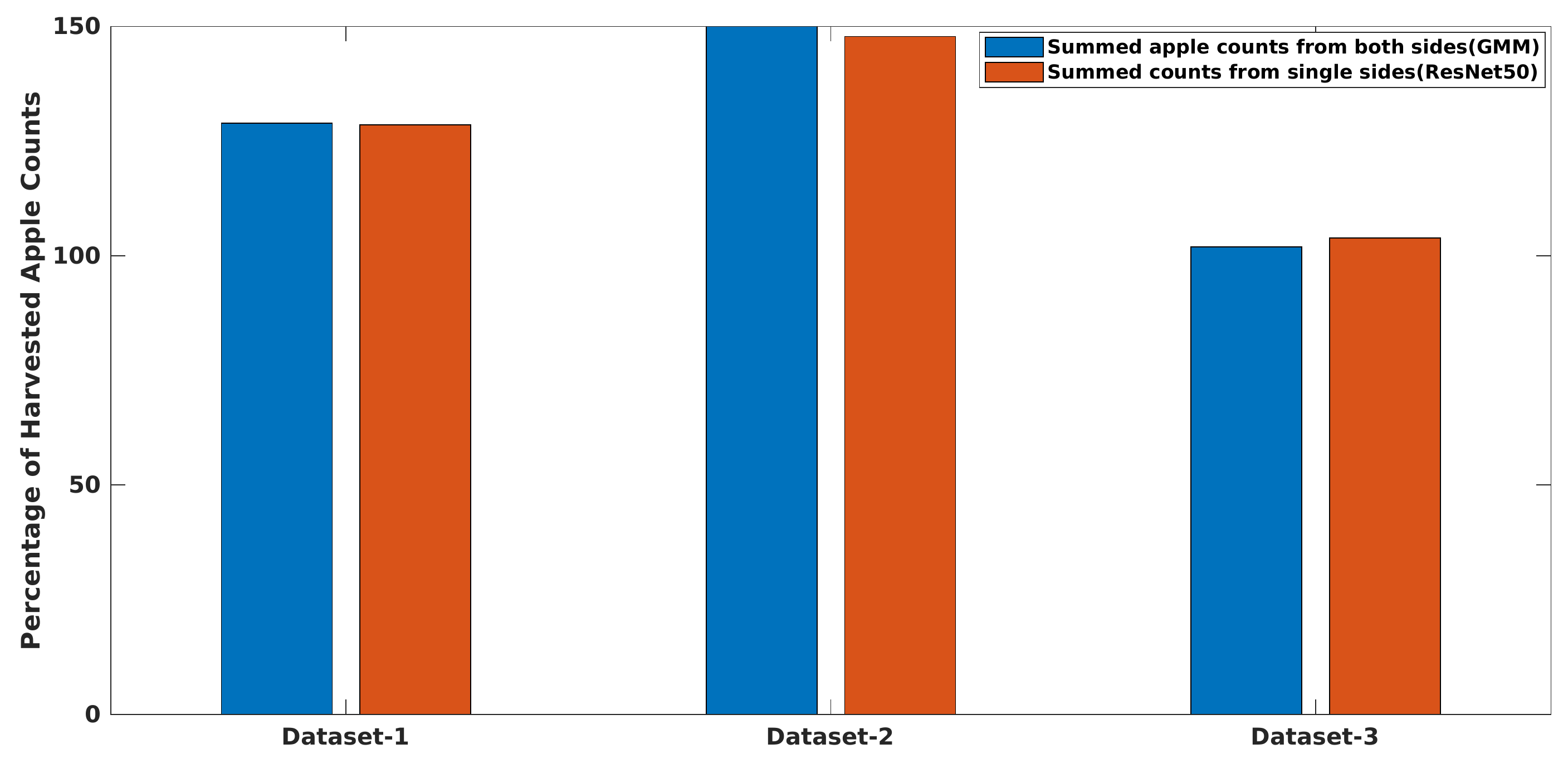}\label{fig:labelgt}}
	\quad
	\centering
	\subfloat[][Merged fruit counts from both sides]{\includegraphics[width=0.44\textwidth]{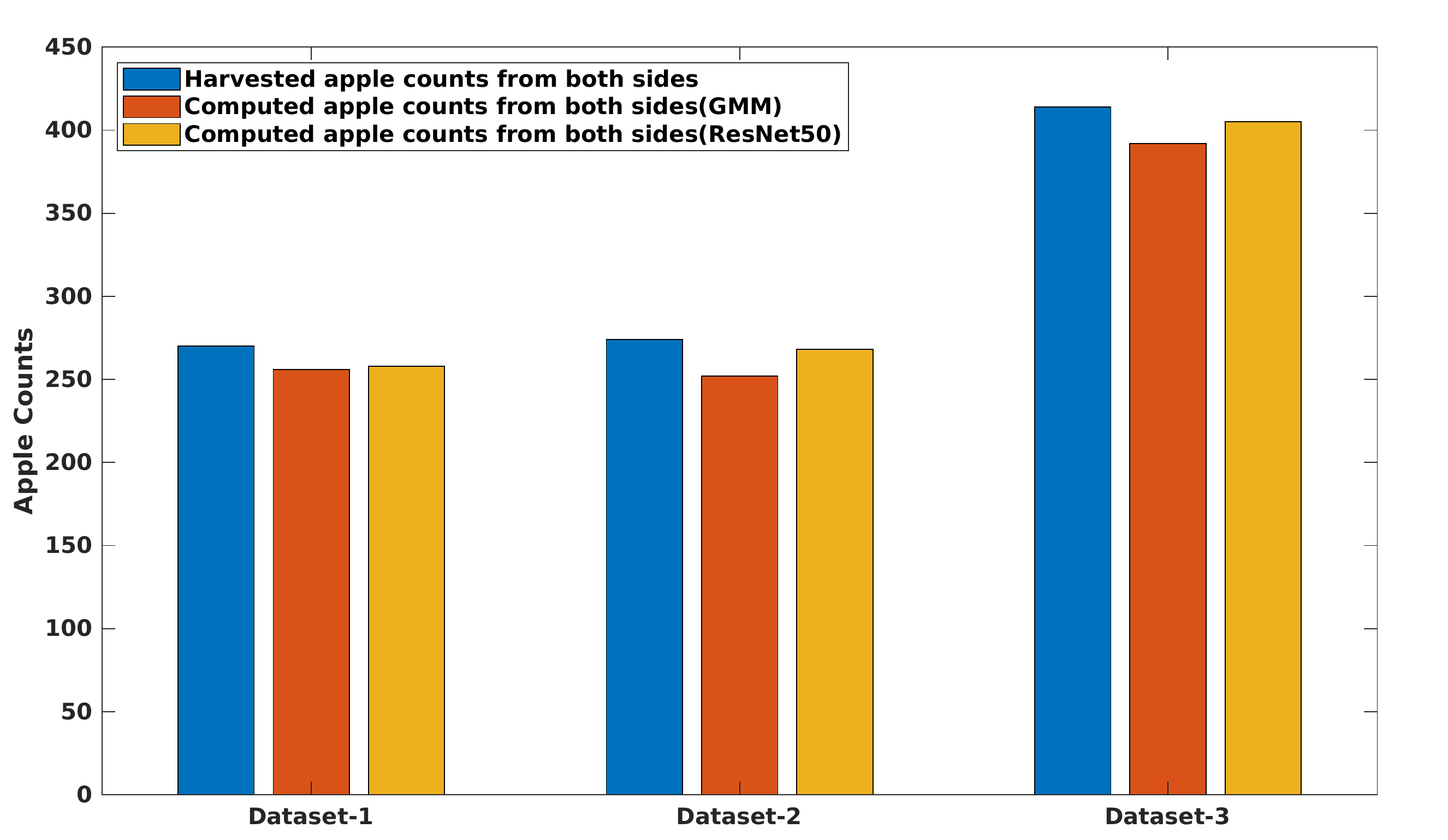}\label{fig:yieldsum}}
	\caption{Fruit yield estimation. (a) Independently summed fruit counts from individual sides lead to inconsistent estimates. (b) Merging the fruit counts from both side, we obtain a more consistent estimate from both GMM ($91.98\%\sim94.81\%$) and ResNet50 ($95.56\%\sim97.83\%$) based counting methods.}
	\label{fig:bothsideyield}
\end{figure}

We conduct these experiments solely with the GMM detection method since the U-Net and FRCNN approach did not show satisfactory detection rates. We evaluate which of the discussed counting methods (GMM/ResNet50) is more useful for yield estimation and quantify the overall yield estimation accuracy of the entire system. Additionally, we demonstrate that tracking fruits visible from both sides leads to more consistent results. We use the three test datasets (Dataset 1, Dataset 2, and Dataset 3); where videos from both sides of a row were collected. The computed yield estimates are compared to the harvested ground truth counts.
Figure~\ref{fig:yieldsum} and Table~\ref{tab:yield} show the shortcomings of adding fruit counts from individual sides independently. These yield estimates vary considerably across datasets for both, the GMM($101.93\%\sim150\%$) and ResNet50 ($103.86\%\sim147.81\%$). However, this method of summing up counts from both sides achieves the lowest error rate $(1.93\%)$ on Dataset 3. Although it does well on this one dataset, it over-counts by up to $50\%$ on others, which makes finding a consistent mapping of these counts to the actual yield tedious and error-prone.

\begin{table}[htb]
	\caption{Summary of yield results in terms of fruit counts (FCs).} \label{tab:yield}
	\begin{center}
		\begin{tabular}{|c|c|c|c|c|c|}
			\hline
			\multirow{2}{*}{Datasets} & \multirow{2}{*}{Harvested FCs} & \multicolumn{2}{|c|}{Merged FCs from both sides} & \multicolumn{2}{|c|}{Sum of FCs from single sides}\\
			\cline{3-6}
			& & GMM & ResNet50 & GMM & ResNet50\\
			\hline
			Dataset-1 & $270$ & $256$ ($94.81\%$) & \textbf{258 (95.56\%)} & $348$ ($128.89\%$) & $347$ ($128.52\%$)\\
			\hline
			Dataset-2 & $274$ & $252$ ($91.98\%$) & \textbf{268 (97.81\%)} & $411$ ($150\%$) & $405$ ($147.81\%$)\\
			\hline
			Dataset-3 & $414$ & $392$ ($94.68\%$) & $405 (97.83\%)$ & 
			\textbf{422 (101.93\%)} & $430$ ($103.86\%$)\\
			\hline
		\end{tabular}
	\end{center}
\end{table}

Next, we investigate the performance of the GMM and ResNet50 based counting methods when a consistent geometric representation of both sides of a row is available. As shown in Figure~\ref{fig:bothsideyield} and Table~\ref{tab:yield}, both the GMM ($91.98\%\sim94.81\%$) and ResNet50 ($95.56\%\sim97.83\%$) based counting methods provide more consistent estimates compared to the independently summed fruit counts. ResNet50 achieves better performance than the GMM for all of the datasets, with errors of $2.17\% - 4.44\%$ compared to harvested ground truth. These results are even more impressive if we consider that our system has counted only visible apples. The camera does not see any apples that are contained within the tree foliage, and therefore does not count them.

\subsection{Failure Cases}
In Section~\ref{sec:qualitative} we presented some qualitative examples. We now analyze the most common failure cases in more detail and offer insights into how these can be overcome in the future.

\subsubsection{Detection}
\begin{figure}[htb]
	\centering
	\includegraphics[width=\textwidth]{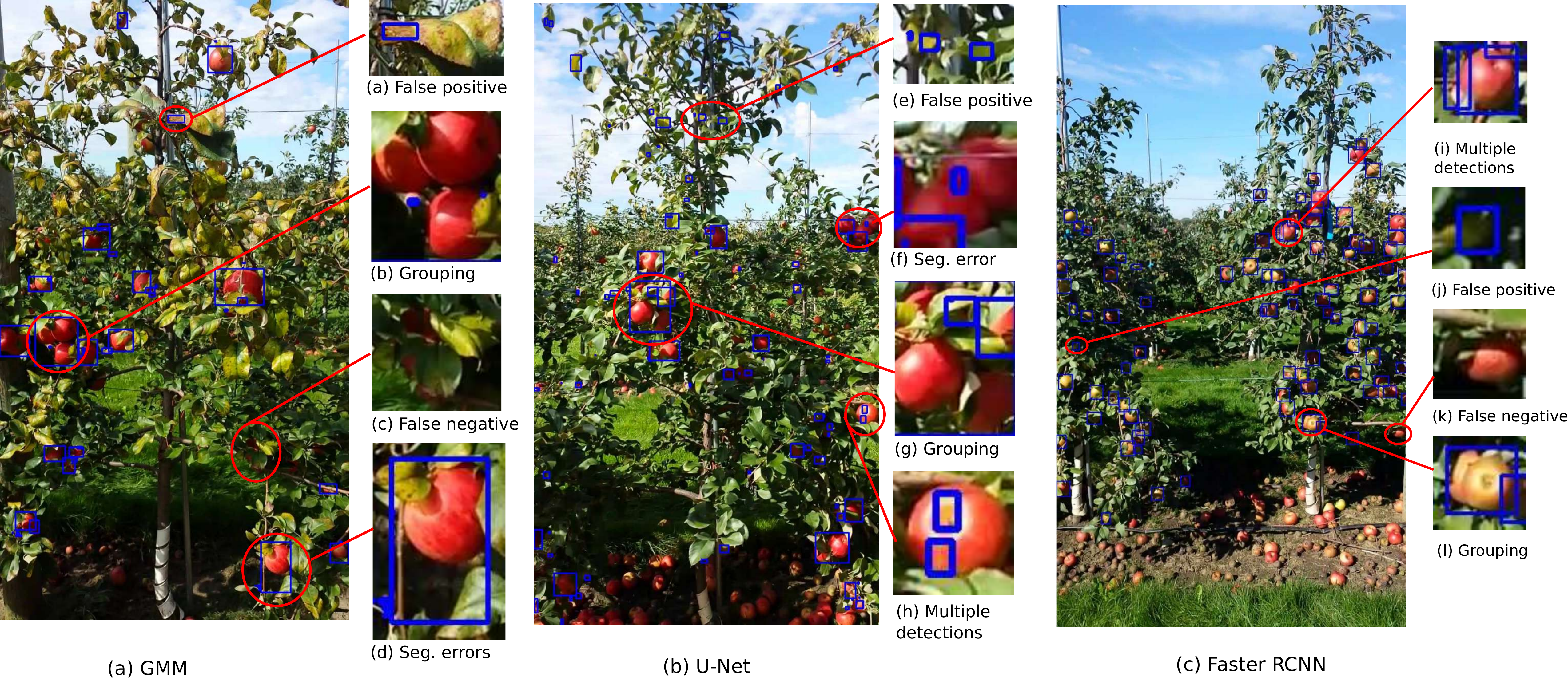}
	\caption{Common failure cases of the three evaluated detection methods}
	\label{fig:failuregmm-unet}
\end{figure}

Some common failure cases of the detection stage are shown in Figure~{\ref{fig:failuregmm-unet}}.

The three methods have similar causes of errors, namely grouping of object instances, false positive detections and false negatives. In addition to these cases, deep learning based methods also split single objects into multiple detections. For the U-Net and the GMM detection methods, the additional network in the counting stage provides the means to reject false positives in $\sim 85\%$ of the cases. The FRCNN method does not contain an additional network for counting. However, the FRCNN could be changed to classifying instances into cluster counts instead of fruit/background. Such an approach will solve the problem of grouped instances and can reject false positives. However, doing so is not straightforward. Future research would have to determine how to merge overlapping predictions and set up an adequate training procedure.

The problem of false negatives is a more challenging one. It occurs in all three detection methods, but for different reasons. In the GMM-based detection method, the number of clusters to over-segment the images are chosen by the user a priori. If this threshold is too low, the model lacks the representative power to disambiguate among different object categories. If the threshold is too high, developing a training model become tedious, as many color clusters are required to capture all the fruits. In some cases, the fruits might not be distinguishable by color at all.
In the case of the U-Net and the FRCNN, this phenomenon is in part due to a lack of training data. The number of false positives for both of these methods highest on Dataset 1. Acquired in late September, the leaves in this dataset were turning yellow. The change of color impacts the networks' performance due to a lack of similar examples in the training set. An additional reason for false negatives in the FRCNN approach is the usage of Non-Maximum Suppression (NMS). Since NMS uses static thresholds, the network is prone to filter out overlapping true positives. While NMS is the de-facto standard algorithm to reject overlapping instances, it hurts performance when we try to detect individual instances of grouped objects.

\subsubsection{Counting}
\begin{figure}[ht]
	\centering
	\subfloat[]{{\includegraphics[width=0.15\columnwidth]{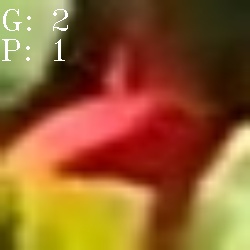}\label{5a} }}%
	\quad
	\subfloat[]{{\includegraphics[width=0.15\columnwidth]{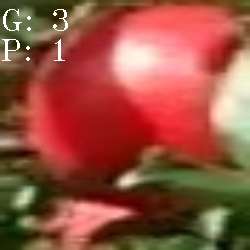}\label{5b} }}%
	\quad
	\subfloat[]{{\includegraphics[width=0.15\columnwidth]{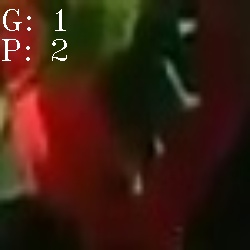}\label{5c} }}%
	\\
	
	\subfloat[]{{\includegraphics[width=0.15\columnwidth]{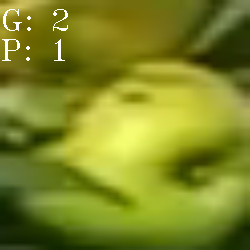}\label{5d} }}%
	\quad
	\subfloat[]{{\includegraphics[width=0.15\columnwidth]{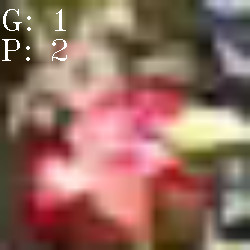}\label{5e} }}%
	\quad
	\subfloat[]{{\includegraphics[width=0.15\columnwidth]{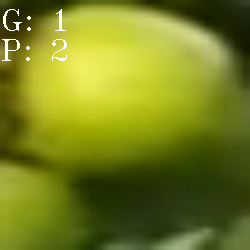}\label{5f} }}%
	\quad
	\caption{Example failure cases of the neural network based counting method}%
	\label{fig:5}%
\end{figure}

The deep learning based counting approach, although achieving overall $90.5\%$ accuracy, contains a few failure cases.  Compared to our experiments in~\cite{hani_apple_2018}, we removed images where apples were lying on the ground from the test set. Removing such fruits is warranted since we use segmentation mask, obtained from 3D reconstruction, to remove apples on the ground or background trees.  Even with these changes and with using a deeper network we cannot remove all failure cases. Figure~\ref{fig:5} shows, that errors often happen when fruits are partially visible. This problem can be only be avoided if the detection method returns patches that show only whole fruits. Since fruits are often occluded, this scenario is not realistic. 
A second problem can be observed in Figure~\ref{5b}. Here the label was wrongly annotated (it should be 2 instead of three). Additionally, the fruits in this image have substantial overlap. When annotating these images, individual labels are often inconsistent. Human labeling errors are present in most datasets, especially if the annotated scenes are cluttered. To avoid them, we annotate the datasets by multiple people and choose the median annotation. However, this increases the human labeling effort drastically.

\section{Conclusion and Future Work}\label{sec:conclusion}
In this paper, we presented new fruit detection and counting methods and evaluate them for the task of yield estimation. One challenge apparent in the literature is that each proposed method uses a separate dataset for testing. To alleviate this problem, we performed a comparative study of different fruit detection and counting methods. This work is the first one to conduct such a comparison on the same datasets. For fruit detection, the semi-supervised clustering technique, based on Gaussian Mixture Model (GMM) achieved the highest $F_1$-score in six out of seven datasets. Our segmentation approach based on U-Net performed reasonably well, but the reimplemented Faster R-CNN suffered from poor precision. For fruit counting, the Convolutional Neural Network (CNN) approach was more accurate for both, single image datasets and yield estimation. Additionally, together with our recent work~\cite{roy_registering_2018,dong_semantic_2018}, we presented a complete system for yield estimation. The classical segmentation method combined with the CNN based counting approach achieved yield accuracies ranging from $95.56\%-97.83\%$ compared to the harvested ground truth.

Our results provide quantitative insights into how much we gain in terms of performance with deep learning approaches. For fruit counting, the neural network provides more accurate and robust results. When fruits are distinguishable by color, the evaluated classical detection method outperformed both, the U-Net and FRCNN. The U-Net approach performed exceedingly well when the test dataset was similar to the training dataset (for example U-Net on Dataset-4 (front)). However, for fruit detection, many challenges remain. It has hard to offer conclusive insights towards the generalizability of the U-Net and FRCNN approaches based on the limited amount of data we had available for training. We plan to increase the size of the training data in the future, so that further research can give insights into this question.

Obtaining more data involves labeling fruit boundaries in images. However, labeling by skilled laborers is time intensive and costly. In the future, we plan to explore the use of synthetic data as training data, eliminate the painstaking process of labeling. The advantage of this approach is that labels are readily available. The disadvantage is that models naively trained on synthetic data do not typically generalize to real data.
\subsubsection*{Acknowledgements}
This work was supported by the USDA NIFA MIN-98-G02. The authors acknowledge the Minnesota Supercomputing Institute (MSI) at the University of Minnesota for providing resources that contributed to the research results reported within this paper~\url{http://www.msi.umn.edu}.

\bibliographystyle{apalike}
\bibliography{jfr2018refs}

\end{document}